\documentclass[sigconf]{acmart}

\renewcommand\footnotetextcopyrightpermission[1]{}

\newcommand{\sys}{\textsc{CacheScout}\xspace}
\newcommand{\kvflow}{KVFlow\xspace}
\newcommand{\continuum}{Continuum\xspace}

\newcommand{\sglang}{SGLang\xspace}

\newcommand{\rui}[1]{{\color{magenta}{(Rui: #1)}}}

\usepackage{amsthm}
\usepackage{algorithm}
\usepackage{algpseudocode}
\usepackage{xspace}
\usepackage{siunitx}
\usepackage{tikz}
\usepackage[normalem]{ulem}
\usepackage{pifont}
\usepackage{array}
\usepackage{enumitem}
\usepackage{booktabs}
\usepackage{comment}
\usepackage[skip=1pt]{subcaption}
\usepackage[dvipsnames]{xcolor}
\usepackage{makecell}

\newcommand{\cmark}{\ding{51}}
\newcommand{\xmark}{\ding{55}}

\begin{document}

\title{Learning Agent Execution for KV-Cache Management in Agentic Serving}
\subtitle{Research Full}

\author{Rui Zhang\textsuperscript{1}, Chaeeun Kim\textsuperscript{1}, Shaoting Feng\textsuperscript{2}, Kuntai Du\textsuperscript{3}, Yuhan Liu\textsuperscript{3}, Yi Zhong\textsuperscript{1}, Cheng-Wei Ching\textsuperscript{1}, Junchen Jiang\textsuperscript{3}, Liting Hu\textsuperscript{1}}
\affiliation{%
  \institution{\textsuperscript{1}University of California, Santa Cruz \quad
               \textsuperscript{2}University of Washington \quad
               \textsuperscript{3}University of Chicago}
  \country{}}

\renewcommand{\shortauthors}{Zhang et al.}

\begin{abstract}
  Multi-agent LLM systems have emerged as an important deployment paradigm for AI services, where each user request is decomposed into a sequence of specialized agents. Across these workflows, every agent repeatedly executes a fixed context consisting of system prompts, tool definitions, and few-shot examples, creating substantial opportunities for KV-cache reuse. Existing LLM serving systems, however, manage KV-cache reactively using prefix caching and recency-based replacement, causing reusable agent contexts to be evicted before their next invocation and forcing repeated recomputation.
We present \sys, an agent-aware KV-cache runtime layer for multi-agent LLM serving. The key insight is that future KV-cache reuse is governed by agent execution semantics rather than cache recency alone. \sys captures these semantics by learning agent execution transitions online, without requiring predefined workflow graphs or offline training, and uses the learned execution model to guide both cache eviction and proactive prefetching while leaving the serving critical path unchanged. We implement \sys on top of vLLM. Across representative real-world multi-agent workloads, \sys improves KV-cache hit rate by 10--18 percentage points, reduces mean TTFT by 18--45\%, lowers mean per-turn latency by 29--38\%, and increases peak throughput by up to 57\%. These benefits also generalize to larger models, reducing TTFT by up to 54\% while sustaining 37\% higher throughput.
\end{abstract}

\keywords{
LLM serving, multi-agent systems, KV cache, cache management, prefix caching, machine learning systems
}

\maketitle

\section{Introduction}
\label{sec:intro}

\begin{figure}[t]
\centering
\includegraphics[width=0.95\columnwidth]{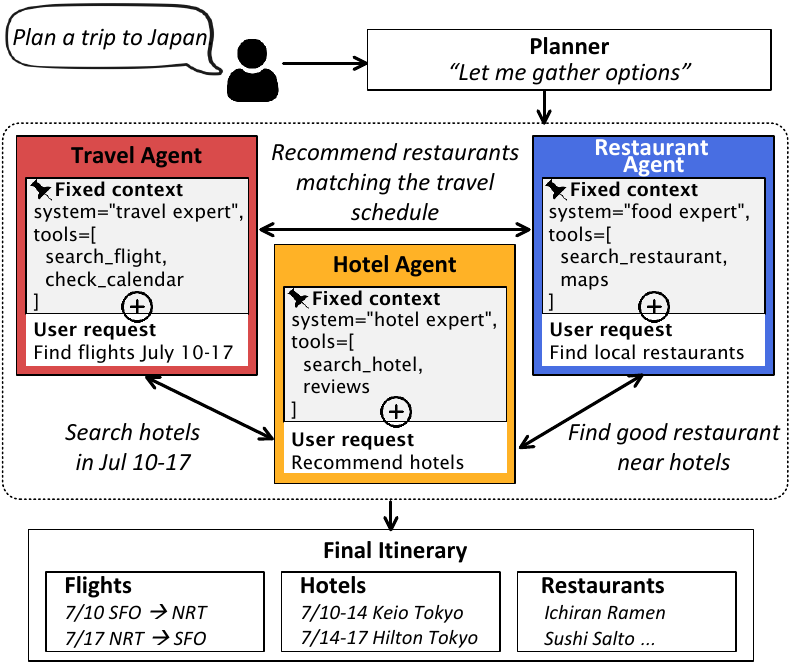}
\caption{Multi-agent travel planning use case.}
\label{fig:intro_usecase}
\vspace{-0.1in}
\end{figure}

Multi-agent LLM systems have rapidly become the dominant deployment paradigm for production AI services, powering applications such as software engineering assistants~\cite{yang2024sweagent, jimenez2024swebench, zhang2024autocoderover}, enterprise knowledge assistants~\cite{lewis2020retrieval, gao2023rag, syarubany2025pentarag}, and deep research agents~\cite{zhang2025deepresearch, huang2025deep, du2025deepresearch}. Instead of relying on a single LLM call, these applications decompose each user request into subtasks handled by a sequence of specialized agents that collaborate to complete complex tasks. A planner coordinates the workflow, domain-specific agents perform specialized reasoning or tool use, and their results are integrated into a final response.

\begin{figure}[t]
\centering
\vbox to 0pt{\vss
\hspace{0.2in}
\includegraphics[width=0.7\columnwidth]{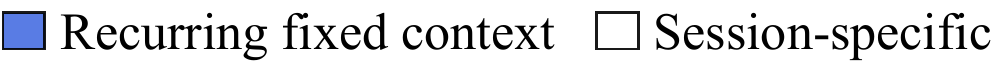}
\vss}
\vspace{0.15in}
\hspace{-0.4in}
\includegraphics[width=0.68\columnwidth]{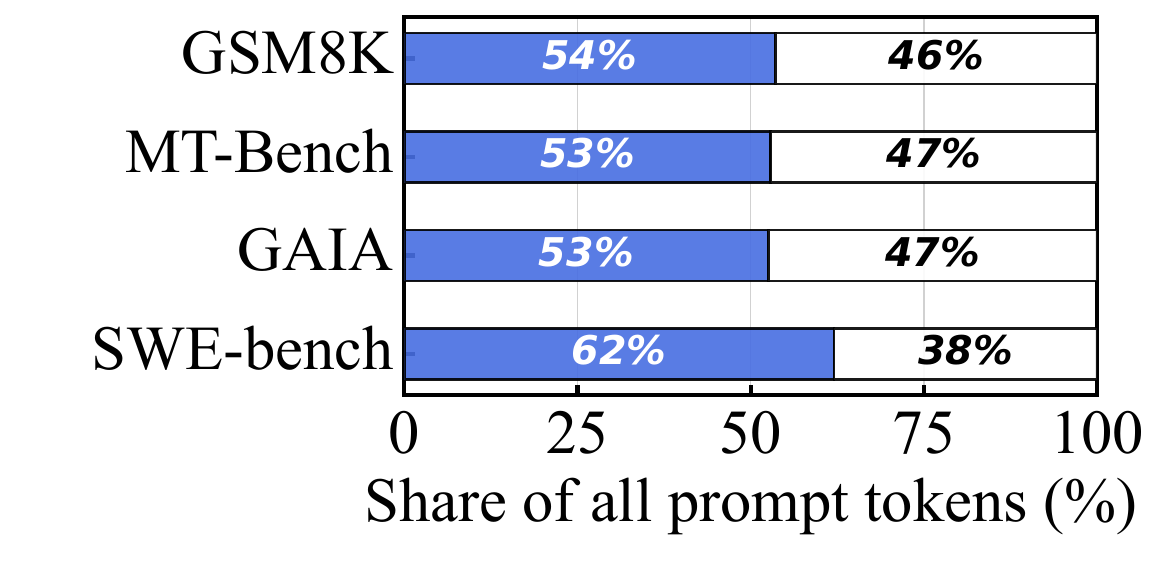}
\vspace{-0.1in}
\caption{Share of prompt tokens occupied by recurring fixed context (system prompts, tool definitions, skills, and examples) in four representative multi-agent workloads: GSM8K~\cite{cobbe2021training}, MT-Bench~\cite{zheng2023judging}, GAIA~\cite{mialon2024gaia}, and SWE-bench~\cite{jimenez2024swebench}.}
\label{fig:anchor-share}
\end{figure}

As shown in Figure~\ref{fig:intro_usecase}, consider a user asking an AI assistant to plan a trip to Japan. The planner first asks a Travel Agent to search for flights, then a Hotel Agent to recommend places to stay, and finally a Restaurant Agent to find local restaurants before putting everything together. Every time the Travel Agent is invoked, it begins with the same fixed context, including a system prompt such as \texttt{You are a travel expert. Find the best flight options based on the user's budget and schedule.} together with the same flight-search tool definitions. The Hotel Agent likewise begins with \texttt{You are a hotel recommendation expert...} together with the same hotel-search tools. The task-specific user request is appended after this fixed context. Since this fixed context rarely changes, the KV-cache generated for it can be reused whenever the same agent is invoked, even across different user sessions. As shown in Figure~\ref{fig:anchor-share}, this recurring fixed context accounts for 53--62\% of all prompt tokens across the four multi-agent workloads evaluated in this paper, making it a substantial opportunity for KV-cache reuse.

However, today's LLM serving systems, such as vLLM~\cite{kwon2023vllm}, SGLang~\cite{zheng2024sglang}, and TensorRT-LLM~\cite{nvidia2023tensorrtllm}, do not fully exploit this opportunity because their cache management is largely reactive, relying on prefix caching and recency-based cache replacement. Returning to our example, after the Travel Agent finishes searching for flights, the planner continues with the Hotel Agent and the Restaurant Agent. As their KV blocks gradually fill the GPU cache, the KV-cache corresponding to the Travel Agent's fixed context is eventually evicted because it has not been accessed recently. Later, after reviewing the hotel options, the planner decides to compare another flight itinerary and invokes the Travel Agent again. Since its reusable fixed context has already been evicted, the serving system must recompute the same prefix from scratch. This pattern repeats throughout a multi-agent workflow. Although a reusable agent anchor may remain idle while other agents execute, it is likely to be reused when the same agent is invoked again. Recency-based cache policies cannot distinguish this temporary idle period from genuinely cold data, leading them to evict precisely the KV blocks that are most valuable to future execution.

\noindentparagraph{\textbf{The problem.}} Closing the gap requires the cache to retain KV blocks that will be used again, instead of simply keeping the blocks that were used most recently. This is much harder than it appears. Today's serving engines, such as vLLM~\cite{kwon2023vllm} and SGLang~\cite{zheng2024sglang}, manage KV-caches as anonymous blocks indexed by content hashes. They know which block was accessed, but they do not know \textbf{who} used it. For example, the engine cannot tell whether a block belongs to the Travel Agent, the Hotel Agent, or the Restaurant Agent. As a result, it cannot tell whether a block is likely to be reused in the near future or can be safely evicted.

Even if the cache knew which agent produced each block, deciding what comes next is still difficult. Modern agent frameworks, such as AutoGen~\cite{wu2024autogen} and LangGraph~\cite{langgraph}, do not follow a fixed workflow. Instead, the next agent is chosen by the LLM at runtime. The planner may call the Travel Agent again, switch to another agent, or finish the task altogether. Therefore, future reuse cannot be determined from a predefined workflow. Recent systems sidestep this difficulty by requiring the workflow to be known in advance, either as a declared execution graph or developer-provided annotations~\cite{pan2026kvflow, lin2024parrot, guo2026saga}, or by pinning the current session's KV-cache across tool calls~\cite{li2025continuum, abhyankar2024infercept}. Neither approach helps when the next agent is decided dynamically at runtime.

Finally, any prediction must be extremely fast. Cache eviction happens whenever GPU memory becomes full and is part of the serving hot path. Learning-based cache replacement policies from the storage literature~\cite{song2020learning, vietri2018lecar, yang2023glcache, beckmann2018lhd} improve prediction quality through feature extraction and model inference, but spending milliseconds to make a better cache decision would easily cost more than the prefill computation it saves.

\noindentparagraph{\textbf{Our solution.}} We address this problem by introducing \sys, a lightweight KV-cache runtime layer that bridges the gap between agent frameworks and serving engines. Instead of managing KV blocks solely based on past accesses, \sys observes agent execution, learns how agents are dispatched at runtime, predicts which agent is likely to execute next, and uses this information to guide cache eviction and prefetching. In this way, cache management becomes aware of future agent reuse rather than merely reacting to previous accesses.

\sys realizes this design through three techniques. (1) \textbf{Transition learner.} It models agent execution as an online first-order Markov chain. For each request, it identifies the agent from the prompt-prefix fingerprint and updates a single transition counter. No offline training or predefined workflow graph is required (Sec.~\ref{sec:design:learner}). (2) \textbf{Survival-probability scorer.} It estimates the probability that each cached anchor will be reused from the learned transition matrix and combines this probability with recency and reconstruction cost to rank eviction candidates. As a result, reusable anchors are retained while execution remains predictable, and the policy naturally falls back toward LRU when it is not (Sec.~\ref{sec:design:scorer}). (3) \textbf{Background prefetch.} It predicts the next agent from the same transition matrix and warms the corresponding anchor between requests, leaving the serving critical path unchanged (Sec.~\ref{sec:design:prefetch}).
Together, these techniques transform the KV-cache from a passive, LRU-based component into an active, agent-aware runtime.



\noindentparagraph{\textbf{Summary of results.}}
We implement \sys{} as a lightweight KV-cache runtime layer on top of vLLM and compare it against vanilla vLLM and Continuum. We evaluate \sys{} on four representative real-world multi-agent workloads using a six-agent supervisor framework with Llama-3.1-8B-Instruct. \sys{} improves KV-cache hit rate by 10--18 percentage points, reaching 81--85\% across all workloads, reduces mean TTFT by 18--45\% and mean per-turn latency by 29--38\%, and delivers 19--57\% higher peak throughput. Under the same latency budget, it sustains $1.7$--$12\times$ the load of vanilla vLLM. These gains also extend to larger models. On Qwen3-235B-A22B, a 235B-parameter mixture-of-experts model, \sys{} reduces mean TTFT by 33--54\% on SWE-bench and delivers 37\% higher throughput. We also show that \sys{} remains robust across different cache capacities and workload conditions while incurring negligible runtime overhead.

\noindentparagraph{\textbf{Contributions.}} 
This paper makes the following contributions.
\begin{itemize}[leftmargin=*, itemsep=0.4ex, topsep=0.5ex]
    \item We identify a new challenge in multi-agent LLM serving: existing prefix caches are reactive and cannot exploit future agent reuse.
    \item We propose \sys, a lightweight KV-cache runtime layer that bridges agent frameworks and serving engines, enabling proactive, agent-aware cache management.
    \item We design and implement \sys with three techniques: online transition learning, survival-probability-based eviction, and asynchronous predictive prefetching, while keeping runtime state and overhead low.
    \item We implement \sys on top of vLLM and show significant improvements in cache hit rate, TTFT, and throughput across representative multi-agent workloads.
\end{itemize}

\section{Motivation and Challenges}
\label{sec:motivation}


\subsection{Background}
\noindentparagraph{\textbf{Multi-agent workflows.}}
Modern LLM applications increasingly organize complex tasks as multi-agent workflows. Instead of relying on a single model invocation, a planner decomposes a user request into a sequence of specialized agents, each responsible for a particular task such as planning, retrieval, coding, or tool use. These agents may follow a predefined execution graph or be dynamically selected by an LLM at runtime. Frameworks such as AutoGen~\cite{wu2024autogen}, LangGraph~\cite{langgraph}, CrewAI~\cite{crewai}, and OpenAI Swarm~\cite{openai2024swarm} have made this execution model increasingly common in production systems.

\noindentparagraph{\textbf{Agent anchor.}}
Although different agents perform different tasks, every invocation of the same agent begins with nearly identical prompt prefixes. These prefixes typically consist of a system prompt describing the agent's role, tool definitions specifying available APIs, skill descriptions, and optionally few-shot examples. Throughout this paper, we refer to this reusable prompt prefix as the agent anchor. Since an agent anchor rarely changes across invocations or user sessions, the KV-cache generated for it can be reused whenever the same agent is executed again. 

\noindentparagraph{\textbf{Prefix caching.}}
Modern serving systems, including vLLM~\cite{kwon2023vllm}, SGLang~\cite{zheng2024sglang}, TensorRT-LLM~\cite{nvidia2023tensorrtllm}, LMCache~\cite{liu2025lmcache}, and NanoFlow~\cite{zhu2024nanoflow}, reduce redundant prefill computation through prefix caching. During the prefill stage, the model computes KV tensors for every prompt token. If a later request shares an identical prompt prefix, these KV tensors can be reused instead of recomputing the entire prefix. Existing systems differ mainly in how they identify identical prefixes. For example, vLLM partitions prompts into fixed-size blocks (16 tokens by default) and identifies reusable KV blocks through block hashing, whereas SGLang uses RadixAttention, which organizes prompt prefixes in a radix tree to enable token-level prefix matching. Despite these implementation differences, all existing systems identify reusable KV-caches solely according to prompt content.

\noindentparagraph{\textbf{Reactive cache management.}}
Existing prefix caches are fundamentally reactive. A KV block becomes reusable only after it has already been requested and inserted into the cache. Likewise, cache replacement policies, such as LRU~\cite{mattson1970evaluation}, LFU~\cite{aho1971principles}, ARC~\cite{megiddo2003arc}, and LIRS~\cite{jiang2002lirs}, retain KV blocks solely according to past accesses. The serving engine operates only on content-addressed KV blocks. It neither knows which agent produced a cached block nor whether that agent is likely to be invoked again. As multi-agent workflows become increasingly dynamic, this reactive design leaves many reuse opportunities unexploited.

\begin{figure}[t]
  \centering
  \begin{subfigure}[t]{0.48\columnwidth}
    \centering
    \includegraphics[width=\linewidth]{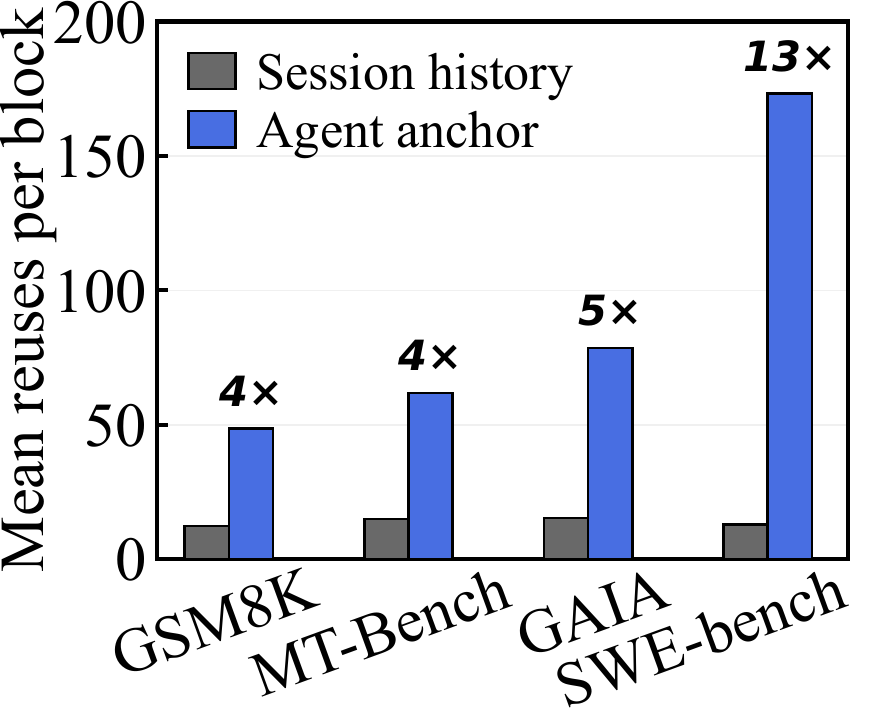}
    \caption{Block reuse counts.}
    \label{fig:anchor-reuse}
  \end{subfigure}
  \hfill
  \begin{subfigure}[t]{0.48\columnwidth}
    \centering
    \includegraphics[width=\linewidth]{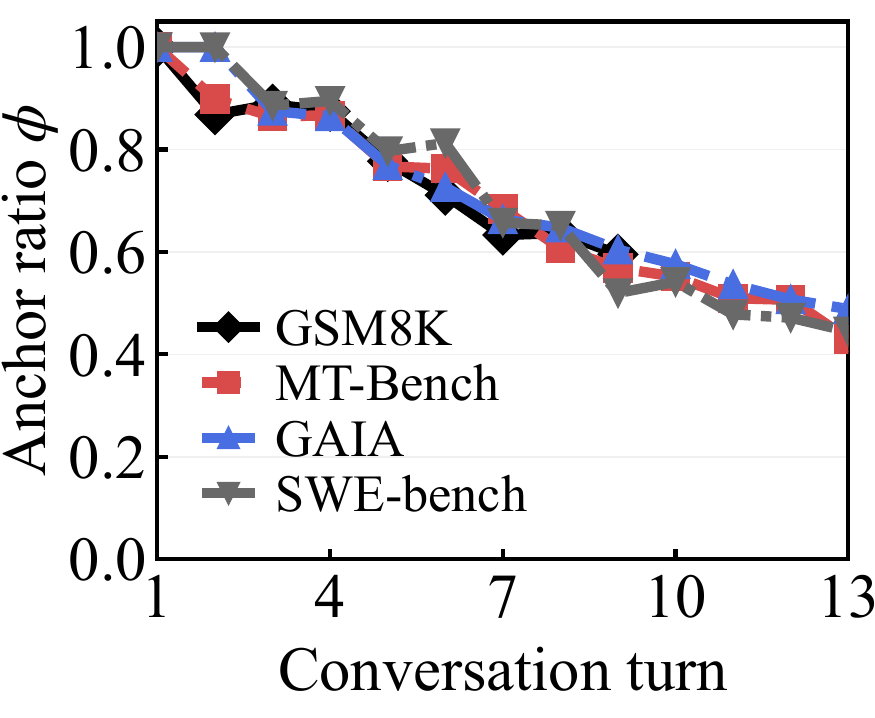}
    \caption{Anchor ratio vs depth.}
    \label{fig:phi-decay}
  \end{subfigure}
  \vspace{-0.1in}
  \caption{Reuse concentration in representative multi-agent workloads. Agent-anchor blocks exhibit substantially higher reuse than session-history blocks, while recurring fixed context remains a significant fraction of each prompt throughout multi-turn interactions.}
  \label{fig:reuse-concentration}
\end{figure}

\subsection{New Insights}
\label{sec:motivation:insights}

The reactive design above leaves the reuse opportunities of multi-agent workloads unexploited. Recovering them requires the cache to act before reuse happens, and whether that is possible depends on the structure of multi-agent execution itself. We therefore ask three questions. First, which cached KV blocks are most valuable to preserve during eviction? Second, is future agent execution predictable enough to act on? Third, can the execution structure be specified before deployment, or must it be learned at runtime? We answer these questions through a measurement study over representative multi-agent workloads, and the resulting insights directly shape the design of \sys.

\subsubsection{Which KV blocks should be protected?}
\mbox{}\\[1ex]

A natural question is whether all cached KV blocks are equally valuable. If future reuse is uniformly distributed across all KV blocks, conventional cache management policies may already be sufficient. On the other hand, if a small subset of KV blocks contributes disproportionately to future reuse, selectively protecting those blocks could significantly improve cache efficiency.

To answer this question, we first measure how often different types of KV blocks are reused. Figure~\ref{fig:anchor-reuse} reports the average number of future reuses per KV block across four representative multi-agent workloads. Agent-anchor blocks are reused 49--173 times on average because every invocation of the same agent accesses the same fixed context, both within and across user sessions. In contrast, session-history blocks belong to a single conversation and are reused only 12--15 times on average. Overall, agent anchors receive 4--13$\times$ more reuse than session-specific history.

Reuse frequency alone, however, does not determine cache value. We therefore quantify how much of each prompt is occupied by reusable agent anchors. We define the \textbf{agent anchor ratio} ($\phi$) as

\begin{equation}
\label{eq:phi}
\phi =
\frac{\text{system prompts}+\text{tool definitions}+\text{skills}+\text{few-shot}}
{\text{total prompt tokens}}.
\end{equation}

Intuitively, $\phi$ measures the fraction of each prompt occupied by the reusable agent anchor rather than session-specific conversation history.

Figure~\ref{fig:phi-decay} plots $\phi$ over the first half of a median session. Although the reusable fraction gradually decreases as conversation history accumulates, agent anchors consistently occupy a substantial portion of every prompt: even a dozen turns into a session, recurring anchor content still accounts for 43--60\% of every prompt.

\noindentparagraph{\textbf{Takeaway}.}
Agent anchors dominate KV-cache reuse for two reasons. They are reused substantially more frequently than session-specific history and occupy a large fraction of every prompt. Protecting anchor blocks therefore provides the greatest opportunity for improving KV-cache reuse.

\begin{figure}[t]
  \centering
  \begin{subfigure}[t]{0.48\columnwidth}
    \centering
    \includegraphics[width=\linewidth]{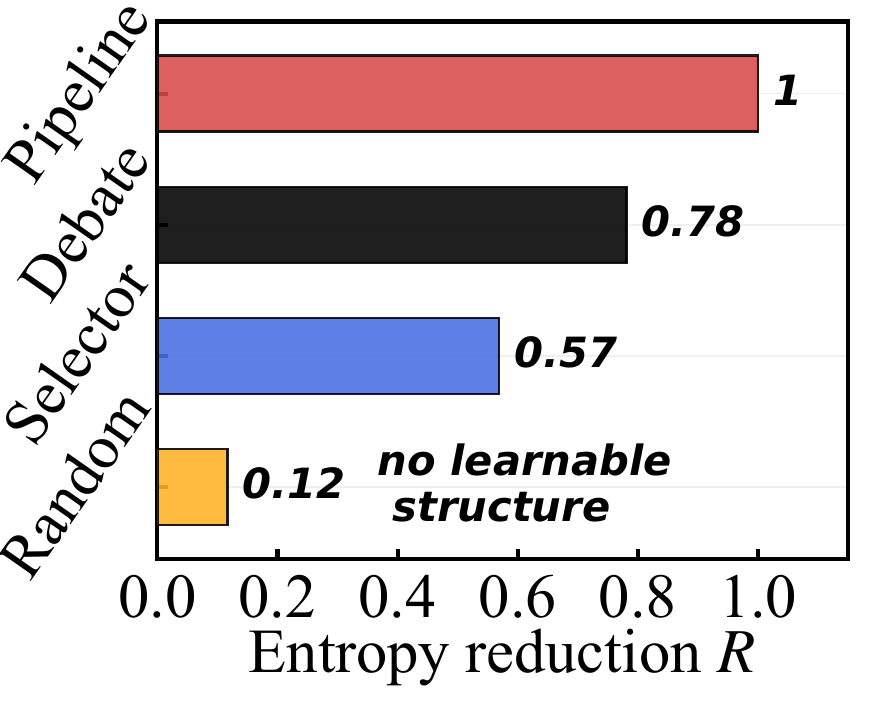}
    \caption{Entropy reduction $R$.}
    \label{fig:topology-r}
  \end{subfigure}
  \hfill
  \begin{subfigure}[t]{0.48\columnwidth}
    \centering
    \includegraphics[width=\linewidth]{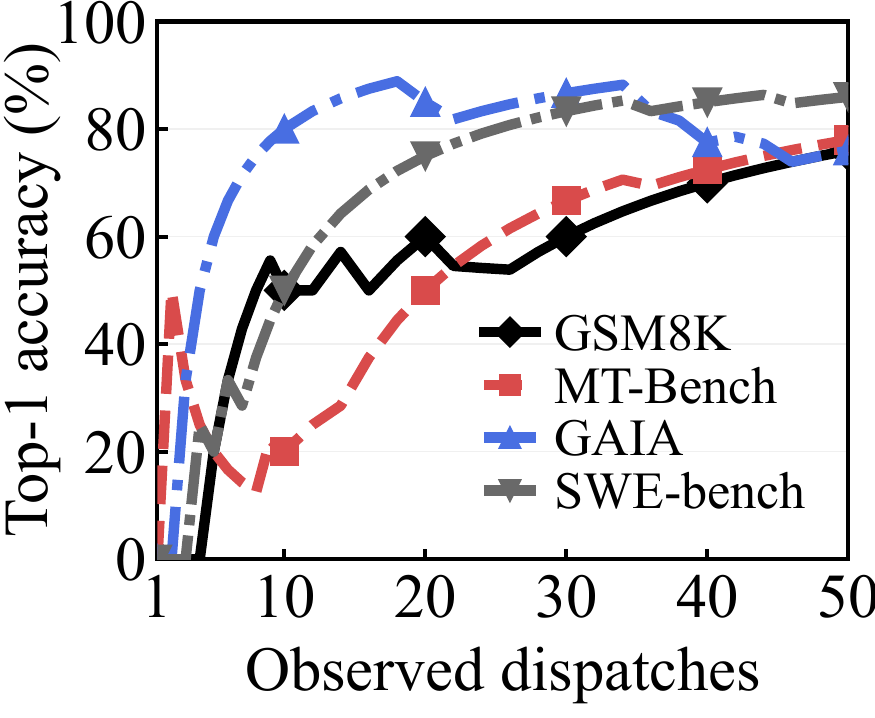}
    \caption{Online prediction accuracy.}
    \label{fig:predict-curve}
  \end{subfigure}
  \vspace{-0.1in}
  \caption{Execution predictability in representative multi-agent workloads. Execution remains structured across diverse coordination topologies, enabling the online transition learner to quickly achieve high next-agent prediction accuracy.}
  \label{fig:predictability}
\end{figure}

\subsubsection{Is future agent execution predictable?}\mbox{}\\[1ex]
Having identified which KV blocks are most valuable to preserve, we next ask whether future agent execution is predictable.

Since proactive cache management only requires predicting the next agent invocation, we quantify predictability using the reduction in uncertainty of the next executed agent. Let $A_t$ denote the agent executed at step $t$. We measure the relative reduction in conditional entropy,

\begin{equation}
\label{eq:predictability}
R \;=\; 1 - \frac{H(A_{t+1} \mid A_t)}{H(A_{t+1})},
\end{equation}
where $R=0$ indicates that the next agent is independent of the current agent, while $R=1$ indicates that the next agent is completely determined by the current one.

\begin{figure*}[t]
  \centering
  \includegraphics[width=0.95\linewidth]{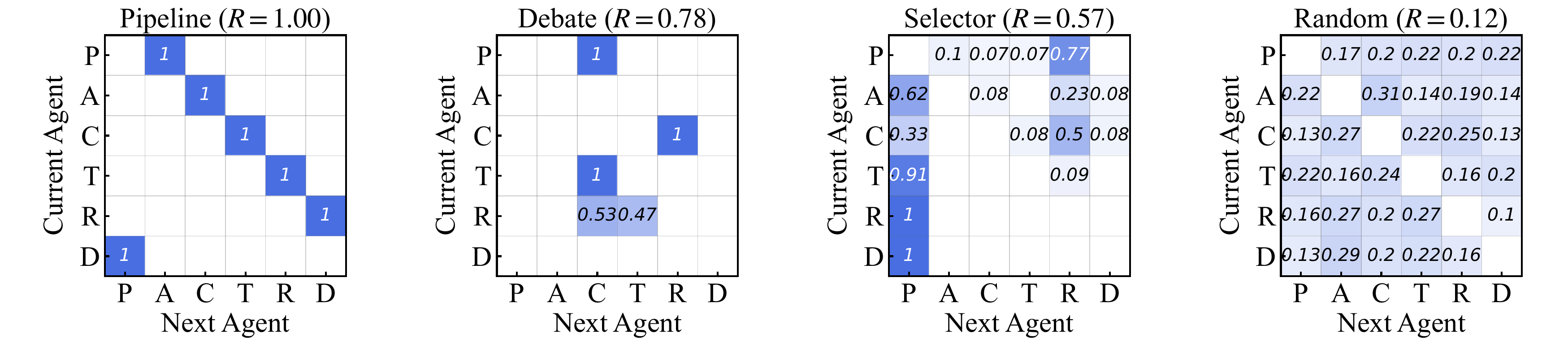}
  \vspace{-0.1in}
  \caption{Transition matrices $P(\mathrm{next}\mid\mathrm{current})$ learned from four coordination topologies built on the same six-agent framework. The same agents produce dramatically different execution structures under different coordination behaviors, so transition models cannot be specified offline and must be learned at runtime.}
  \label{fig:transitions}
\end{figure*}

Figure~\ref{fig:topology-r} reports $R$ for four representative coordination topologies. Pipeline exhibits deterministic execution ($R=1.0$) and the moderated Debate topology remains highly structured ($R=0.78$), while SelectorGroupChat, where the next agent is dynamically selected by an LLM, still achieves $R=0.57$. In other words, conditioning on the current agent eliminates 57\% of the uncertainty in the next agent. Only the Random topology shows little learnable structure ($R=0.12$). Overall, the three practical coordination patterns all exhibit substantial predictability despite making runtime dispatch decisions.

This predictability is also actionable at runtime. Figure~\ref{fig:predict-curve} replays each workload's execution trace through an online first-order transition counter, the same statistic \sys maintains: predicting the next agent as the most frequent observed successor of the current agent reaches 76--86\% top-1 accuracy within 50 observed dispatches, without any offline training.

\noindentparagraph{\textbf{Takeaway}.}
Future agent execution is not random. Although multi-agent workflows are dynamically determined at runtime, they retain strong local transition structure that a lightweight online model can capture within tens of dispatches, making future agent execution predictable enough to drive proactive cache management.

\subsubsection{Can execution structure be specified offline?}\mbox{}\\[1ex]
The previous insight shows that future agent execution is predictable. We finally ask whether this execution structure can be determined before deployment.

To answer this question, we compare the transition matrices of four representative coordination topologies built on the same six-agent framework. Figure~\ref{fig:transitions} shows that execution structure varies dramatically across coordination topologies despite using the same set of agent roles.

Pipeline and Debate are both highly predictable ($R=1.0$ and $R=0.78$), yet their transition matrices are fundamentally different. Pipeline follows a fixed chain through all six agents, whereas Debate concentrates execution on a proposer--challenger exchange between Coder and Reviewer that is periodically adjudicated by a judge. SelectorGroupChat represents the production setting, where the next agent is dynamically selected by an LLM. Although its execution is no longer deterministic ($R=0.57$), transition probability remains concentrated on a few high-probability edges. In contrast, the Random topology distributes probability almost uniformly across all agent pairs ($R=0.12$), leaving little exploitable execution structure.

\noindentparagraph{\textbf{Takeaway}.}
 Execution structure is not fixed by the agent framework itself. Instead, it is determined by runtime coordination behavior. Consequently, transition models cannot be specified offline and must instead be learned online from runtime execution history.

\begin{table*}[t]
\caption{Comparison of representative approaches to KV-cache management for multi-agent serving. \cmark\ supported; \xmark\ not supported.}
\vspace{-0.1in}
\label{tab:comparison}
\footnotesize
\setlength{\tabcolsep}{7pt}
\begin{tabular}{
>{\raggedright\arraybackslash}m{2.9cm}
>{\raggedright\arraybackslash}m{1.9cm}
>{\raggedright\arraybackslash}m{1.9cm}
>{\raggedright\arraybackslash}m{2.0cm}
>{\centering\arraybackslash}m{0.75cm}
>{\centering\arraybackslash}m{0.9cm}
@{\hspace{28pt}}
>{\raggedright\arraybackslash}m{3.5cm}
}
\toprule
\textbf{Approach (Representative Systems)}
&
\textbf{Cache Abstraction}
&
\textbf{Reuse Policy}
&
\textbf{Assumption}
&
\makecell{\textbf{Agent-}\\\textbf{aware}}
&
\makecell{\textbf{Learnable}\\\textbf{Runtime}}
&
\textbf{Limitations for Multi-Agent Serving}
\\
\midrule

\textbf{Reactive serving engines}
\textit{(vLLM~\cite{kwon2023vllm},
SGLang~\cite{zheng2024sglang},
TensorRT-LLM~\cite{nvidia2023tensorrtllm})}
&
Content-addressed KV blocks
&
Prefix matching + recency eviction
&
Recent use predicts reuse
&
\textcolor{BrickRed}{\xmark}
&
\textcolor{BrickRed}{\xmark}
&
Blind to agent identity; recency-based eviction discards anchors with long reuse distances despite their near-certain reuse.
\\
\midrule

\textbf{Static agent-aware KV management}
\textit{(KVFlow~\cite{pan2026kvflow},
Tokencake~\cite{bian2025tokencake},
\continuum~\cite{li2025continuum},
KVCOMM~\cite{ye2026kvcomm})}
&
KV-cache + static workflow state
&
Scheduled prefetch, pinning, or KV adaptation
&
Workflow structure known in advance
&
\textcolor{ForestGreen}{\cmark}
&
\textcolor{BrickRed}{\xmark}
&
Static assumptions break under LLM-selected routing; \continuum\ protects only the current session, and KVCOMM's approximation degrades output quality.
\\
\midrule

\textbf{Learning-based cache replacement}
\textit{(LRB~\cite{song2020learning},
LeCaR~\cite{vietri2018lecar},
GL-Cache~\cite{yang2023glcache},
LHD~\cite{beckmann2018lhd})}
&
Opaque cache objects
&
Learned reuse prediction
&
Access history predicts reuse
&
\textcolor{BrickRed}{\xmark}
&
\textcolor{ForestGreen}{\cmark}
&
Feature extraction and model inference exceed the serving hot-path budget; no execution semantics.
\\
\midrule

\textbf{Agent workflow frameworks}
\textit{(AutoGen~\cite{wu2024autogen},
LangGraph~\cite{langgraph},
CrewAI~\cite{crewai})}
&
\textcolor{BrickRed}{\textbf{N/A}}
&
\textcolor{BrickRed}{\textbf{N/A}}; delegated to the engine
&
Execution semantics are visible
&
\textcolor{ForestGreen}{\cmark}
&
\textcolor{BrickRed}{\xmark}
&
Sees agent execution but has no visibility into or control over the engine's KV-cache.
\\
\midrule

{\textbf{\sys\ (ours)}}
&
\textbf{\textcolor{ForestGreen}{KV-cache + online transition matrix}}
&
\textbf{\textcolor{ForestGreen}{Transition model guided eviction and prefetch}}
&
\textbf{\textcolor{ForestGreen}{Execution structure is learnable}}
&
\textcolor{ForestGreen}{\cmark}
&
\textcolor{ForestGreen}{\cmark}
&
Gains diminish when execution approaches random routing.
\\

\bottomrule
\end{tabular}
\end{table*}

\subsection{Challenges}
\label{sec:motivation:challenges}

Building an agent-aware KV-cache runtime introduces fundamental challenges. Table~\ref{tab:comparison} summarizes the assumptions underlying existing approaches related to KV-cache management and their limitations for multi-agent serving.

\noindentparagraph{\textbf{Challenge \#1: Semantic gap between agents and KV blocks.}}
Existing prefix caches operate on prompt content, not agent semantics. vLLM identifies reusable state through block hashes, while SGLang uses radix-tree prefix matching. These mechanisms can determine whether two prompt prefixes are identical, but they cannot tell which agent produced a KV block, which session it belongs to, or whether the same agent is likely to be invoked again. Prior cache replacement policies, from heuristic approaches such as LRU~\cite{mattson1970evaluation}, LFU~\cite{aho1971principles}, ARC~\cite{megiddo2003arc}, and LIRS~\cite{jiang2002lirs} to learning-based methods such as LRB~\cite{song2020learning}, LeCaR~\cite{vietri2018lecar}, GL-Cache~\cite{yang2023glcache}, and LHD~\cite{beckmann2018lhd}, still make decisions over cache objects. They do not expose the agent-level identity needed to protect reusable agent anchors.

\noindentparagraph{\textbf{Challenge \#2: Dynamic execution without a static graph.}}
Existing workflow-level optimizations assume that future execution can be described by a declared graph, a fixed pipeline, or developer-provided hints. This assumption breaks for modern agent frameworks such as AutoGen SelectorGroupChat and OpenAI Swarm, where the next agent is often selected by an LLM at runtime. As we show in Sec.~\ref{sec:motivation:insights} (Figure~\ref{fig:transitions}), even the same six-agent framework can produce very different transition matrices under different coordination topologies. A cache policy cannot rely on a static workflow graph; it must learn the execution structure from live dispatches.

\noindentparagraph{\textbf{Challenge \#3: Prediction under serving-system constraints.}}
Even when future agent execution is predictable, prediction must operate within the stringent latency budget of a production serving engine. Cache eviction is triggered whenever GPU memory becomes constrained and therefore lies directly on the serving critical path. Existing learning-based cache management techniques often improve prediction quality through complex feature extraction, model inference, or periodic retraining. While effective for traditional storage caches, these techniques introduce overheads that are difficult to justify in LLM serving, where even microseconds of additional latency accumulate across millions of cache operations. The challenge is therefore not simply to predict future reuse accurately, but to do so using lightweight state, microsecond-scale decision latency, and graceful degradation when execution exhibits little predictable structure.
\section{Design}
\label{sec:design}

\begin{figure*}[t]
  \centering
  \includegraphics[width=0.95\linewidth]{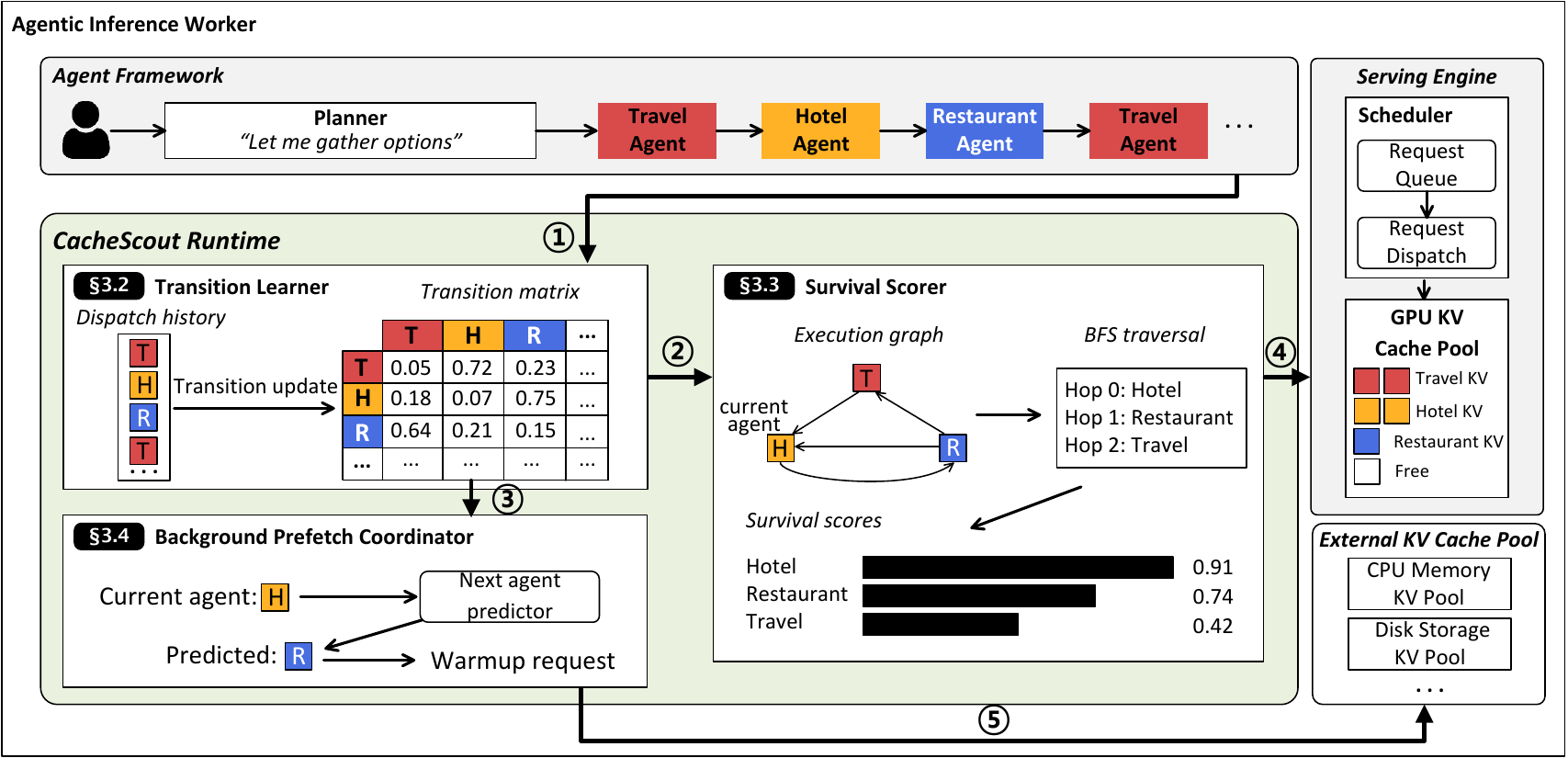}
  \vspace{-0.1in}
  \caption{\sys architecture. \sys sits between the agent framework and the serving engine as a lightweight runtime layer. {\Large\ding{172}} Each agent dispatch is observed by the Transition Learner, which updates the online transition matrix. The learned matrix feeds {\Large\ding{173}} the Survival Scorer and {\Large\ding{174}} the Background Prefetch Coordinator. {\Large\ding{175}} The scorer converts hop distances on the execution graph into survival scores that guide eviction in the GPU KV-cache pool, and {\Large\ding{176}} the coordinator issues a background warmup request for the predicted agent's anchor, reloading it from the external KV pools when present.}
  \label{fig:architecture}
\end{figure*}

\subsection{Runtime Overview}
\label{sec:design:overview}
\sys is not another cache replacement policy. Instead, it rethinks the role of the KV-cache runtime in agentic serving. Existing serving engines efficiently manage KV blocks, but all cache decisions are inherently reactive. They rely solely on past accesses. Our goal is to answer a different question: \textit{Can the serving system prepare for future execution instead of simply reacting to previous execution?}

Our key idea is to introduce an independent \textbf{KV-cache runtime layer} between the agent framework and the serving engine. Rather than embedding agent semantics into the serving engine or pushing cache management into the agent framework, \sys separates cache intelligence into its own runtime. This runtime continuously (1) learns how agents interact, (2) predicts which agents are likely to execute next, and (3) translates these predictions into cache management decisions.

Figure~\ref{fig:architecture} illustrates the overall architecture of \sys. The runtime resides between the agent framework and the serving engine. The agent framework remains responsible for agent scheduling and workflow execution, while the serving engine continues to perform prefix matching, KV allocation, and GPU memory management. \sys introduces no changes to either component. Instead, it exchanges lightweight runtime metadata with both sides, allowing execution information to influence cache management without modifying the underlying serving stack.

\sys realizes this runtime through three cooperating components.
The Transition Learner (Sec.~\ref{sec:design:learner}) is responsible for understanding how a workload executes. As agents interact, it continuously observes runtime dispatches and learns a workload-specific transition model. Instead of assuming a predefined workflow graph, it models execution as an online first-order Markov chain, allowing the runtime to adapt naturally to dynamic routing. 
The Survival Scorer (Sec.~\ref{sec:design:scorer}) answers the next question: which cached state is worth keeping? Rather than asking which block was used most recently, it asks which block is most likely to be reused next. To do so, it estimates the survival probability of each cached agent anchor from the learned transition model and uses this probability to guide cache replacement.
The Background Prefetch Coordinator (Sec.~\ref{sec:design:prefetch}) turns predictions into actions. Once the runtime identifies which agent is likely to execute next, it proactively reconstructs the corresponding high-value KV blocks during idle periods between requests. By moving this work off the serving critical path, \sys improves cache readiness without increasing inference latency.

\subsection{Transition Learner}
\label{sec:design:learner}
The first task of \sys is to infer how a workload executes. This is challenging because modern agent frameworks do not expose a fixed execution graph. The next agent is selected dynamically by LLM reasoning, and the resulting execution structure varies across workloads and evolves over time. 

A natural question is how much execution structure needs to be learned. One option is to model complete execution traces using a sophisticated sequence model. However, cache management does not need to predict an entire workflow. It only needs to anticipate the next scheduling decision before cache replacement occurs. 
Since Sec.~\ref{sec:motivation} shows that local agent transitions already carry substantial predictive signal, \sys learns only the local transition structure needed for cache management.
To realize this idea, the Transition Learner models agent execution as a first-order Markov chain. Let $A_t$ denote the agent executed at step $t$. For every observed transition $A_t = i \rightarrow A_{t+1} = j$, the runtime updates a transition counter $C_{ij} \gets C_{ij} + 1$.
The transition probability from agent $i$ to agent $j$ is then estimated as

\begin{equation}
\label{eq:transition}
P_{ij} = P(A_{t+1} = j \mid A_t = i) = \frac{C_{ij} + \epsilon}{\sum_{k} \left( C_{ik} + \epsilon \right)},
\end{equation}
where $\epsilon$ is a small smoothing constant used to avoid zero-probability transitions early in execution.
Given the current agent $A_t = i$, the predicted distribution over the next agent is simply the $i$-th row of the transition matrix:
\begin{equation}
\label{eq:nextdist}
\mathbf{p}_{t+1} = P_{i,:}.
\end{equation}
The most likely next agent is
\begin{equation}
\label{eq:nextagent}
\hat{A}_{t+1} = \arg\max_{j} P_{ij}.
\end{equation}
Because each dispatch updates only one counter and one row of the transition matrix, the learner has constant update cost and bounded state. It requires no offline training, no workflow annotations, and no replay of historical traces. The result is a lightweight execution model that evolves with the workload and provides the predictive signal needed by the rest of \sys.

\begin{figure}[t]
  \centering
  \includegraphics[width=\columnwidth]{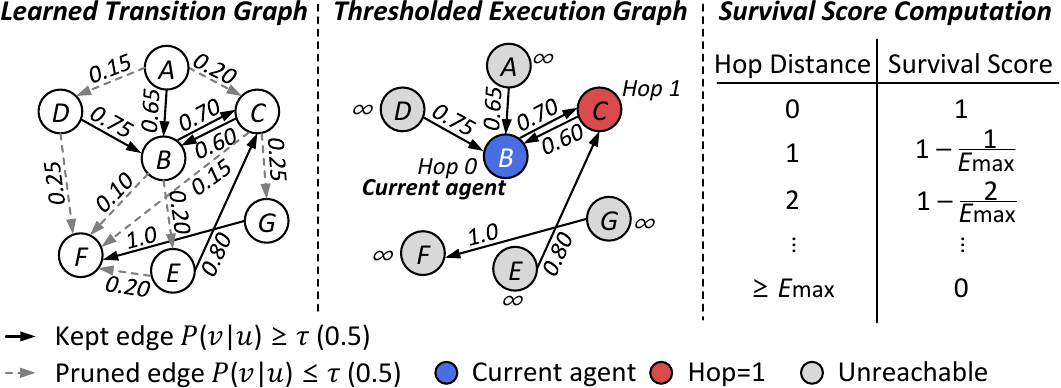}
  \vspace{-0.1in}
  \caption{Survival score computation. \sys converts the learned transition graph into a sparse execution graph by pruning low-confidence transitions, then computes survival scores from BFS hop distances.}
  \label{fig:traversal}
\end{figure}

\subsection{Survival Scorer}
\label{sec:design:scorer}
The second task of \sys is not simply to predict which agent will execute next, but to determine which cached agent anchors are worth preserving. A highly probable agent does not necessarily imply that all of its cached blocks should remain in GPU memory, while an infrequently executed agent may still own reusable anchors that are expensive to recompute. Consequently, transition probability cannot be used directly as an eviction policy.

Ideally, we would rank every cached agent anchor according to its probability of being reused within the next $K$ scheduling steps. We define this quantity as the survival probability.
\begin{equation}
\label{eq:psurv-def}
p_{\mathrm{surv}}(a; a_0, K) = \Pr\left( \exists\, k \in [1, K] : A_{t+k} = a \;\middle|\; A_t = a_0 \right),
\end{equation}
where $a_0$ is the current agent and $a$ denotes a candidate agent anchor. Intuitively, the survival probability measures whether an anchor is likely to remain useful in the near future, making it a natural objective for cache replacement.

The exact survival probability is prohibitively expensive to compute, as it requires aggregating transition probabilities over all execution paths within the prediction horizon. Instead, \sys approximates survival using the learned execution graph. It first converts the transition matrix into a sparse directed graph by retaining only high-confidence transitions,
\begin{equation}
\label{eq:graph}
(a,b) \in E \Leftrightarrow P(b \mid a) \ge \tau,
\end{equation}
where $\tau$ is a confidence threshold.

Figure~\ref{fig:traversal} illustrates the approximation process. Starting from the current agent, the runtime performs a single breadth-first search (BFS) over the sparse execution graph to compute the minimum hop distance $E[a]$ to every reachable agent. Hop distance serves as an efficient proxy for future reuse: anchors expected to be revisited sooner receive higher survival scores than those farther away in the execution graph.
The survival score is then computed as
\begin{equation}
\label{eq:psurv}
\tilde{p}_{\mathrm{surv}}(a) = 1 - \frac{\min\left( E[a], E_{\max} \right)}{E_{\max}},
\end{equation}
where $E_{\max}$ bounds the prediction horizon.

Each cached KV block inherits the survival score of its corresponding agent anchor. To improve robustness, \sys blends the predicted survival score multiplicatively with a recency signal and weights the result by the block's reconstruction cost,
\begin{equation}
\label{eq:score}
\mathrm{Score}(b) = \left( \tilde{p}_{\mathrm{surv}}(a_b) + \delta \right) \cdot \left( e^{-\lambda \cdot \mathrm{age}(b)} + \delta \right) \cdot |b|,
\end{equation}
where $a_b$ is the anchor associated with block $b$, $\mathrm{age}(b)$ counts scheduler steps since the block was last accessed, $\lambda$ is a recency decay rate, and $|b|$ is the number of tokens the block caches. Blocks with lower scores are evicted first. The product can be read as the expected prefill work lost by evicting $b$: the first two factors approximate the probability that the block is reused soon, and $|b|$ is the recompute cost paid if that reuse misses. Between blocks with equal reuse probability, \sys therefore evicts the one that is cheaper to reconstruct. The multiplicative blend lets either signal compensate when the other is weak: under predictable execution, the survival factor protects high-value anchors; when execution becomes less predictable, survival scores flatten and the recency factor dominates, so the policy gracefully degrades toward LRU. The small floor $\delta$ keeps both factors positive so that newly observed blocks, for which the learner has no signal yet, are not immediately evicted at cold start.

\begin{algorithm}[t]
\caption{\sys runtime workflow. Numbered steps correspond to the arrows
in Figure~\ref{fig:architecture}.}
\label{alg:runtime}
\begin{algorithmic}[1]
\Statex \textbf{State:} transition counts $n(\cdot,\cdot)$;
transition matrix $\widehat{P}$; current agent $a_t \gets \bot$;
execution graph $\mathcal{G}$; hop-count table $E[\cdot]$;
block-to-agent map $\textsc{agentOf}[\cdot]$;
last-access table $\textsc{lastAccess}[\cdot]$;
decay rate $\lambda$, floor $\delta$, threshold $R_{\min}$.
\Statex

\Loop
  \State $r \gets$ framework.\textsc{NextRequest}()
        \Comment{agent dispatch}
  \State engine.\textsc{Serve}($r$)
        \Comment{normal serving path}
  \State \Call{BetweenStep}{}
        \Comment{{\Large\ding{174}} off critical path}
\EndLoop

\Statex
\Statex During \textsc{Serve}, the block pool raises two events:
\State \textbf{on} engine.pool.\textsc{BlockTouch}$(b, s_b)$:
\State \quad \Call{ObserveTouch}{$b,s_b$}
        \Comment{{\Large\ding{172}} learn transition}
\State \textbf{on} engine.pool.\textsc{Evict}:
\State \quad \Return $\displaystyle \arg\min_{b \in \mathcal{B}_{\mathrm{evict}}}
        \Call{ScoreBlock}{b}$
        \Comment{{\Large\ding{175}} survival-guided eviction}

\Statex
\Procedure{ObserveTouch}{$b,s_b$}
  \State $a \gets \textsc{AgentId}(s_b)$
  \State $\textsc{agentOf}[b] \gets a$;\;
         $\textsc{lastAccess}[b] \gets \textsc{now}()$
  \If{$a \ne a_t$}
     \If{$a_t \ne \bot$}
        \State $n(a_t,a) \gets n(a_t,a)+1$
        \State update $\widehat{P}$ from $n(\cdot,\cdot)$
     \EndIf
     \State $\mathcal{G} \gets \textsc{ThresholdGraph}(\widehat{P},\tau)$
        \Comment{{\Large\ding{173}} refresh scorer state}
     \State $E[\cdot] \gets \textsc{Bfs}(\mathcal{G},a)$
     \State $a_t \gets a$
  \EndIf
\EndProcedure

\Statex
\Function{ScoreBlock}{$b$}
  \State $a \gets \textsc{agentOf}[b]$
  \State $\mathrm{age} \gets \textsc{now}() - \textsc{lastAccess}[b]$
  \State \Return $(\tilde{p}_{\mathrm{surv}}(a) + \delta) \cdot
        (e^{-\lambda \cdot \mathrm{age}} + \delta) \cdot |b|$
\EndFunction

\Statex
\Procedure{BetweenStep}{}
  \If{$R(\widehat{P}) \ge R_{\min}$}
     \State $a^* \gets \arg\max_a \widehat{P}(a \mid a_t)$
     \State \Call{IssueWarmupRequest}{$a^*$}
        \Comment{{\Large\ding{176}} background warmup}
  \EndIf
\EndProcedure
\end{algorithmic}
\end{algorithm}

\subsection{Background Prefetch Coordinator}
\label{sec:design:prefetch}
The final task of \sys is to proactively prepare reusable KV-cache before it is requested. While the Transition Learner discovers execution structure and the Survival Scorer determines which cached agent anchors should be retained, neither helps when a future anchor has already been evicted or has not yet been constructed. The Background Prefetch Coordinator closes this loop by proactively warming up high-value agent anchors during idle periods. Algorithm~\ref{alg:runtime} summarizes the overall runtime workflow.
A straightforward solution would be to directly insert predicted KV blocks into the cache. However, such a design requires intrusive modifications to the serving engine's memory manager and cache implementation, reducing portability across different inference engines. \sys instead adopts a lightweight approach that reuses the engine's existing prefill pipeline. Rather than manipulating KV blocks explicitly, the coordinator issues a lightweight warmup request containing only the predicted agent's reusable anchor, including its system prompt, tool definitions, and a minimal user prompt. Since this warmup follows the engine's normal execution path, the resulting KV blocks are constructed exactly as they would be during a real agent invocation, requiring no modifications to the attention kernels, memory allocator, or cache implementation.
Given the current agent $a_t$, the coordinator first selects the most likely next agent,
\begin{equation}
\label{eq:prefetch}
a^{*} = \arg\max_{a} P(a \mid a_t),
\end{equation}
and triggers warmup only for the corresponding reusable anchor. Because warmup executes during idle periods between requests, all KV construction occurs off the serving critical path, allowing otherwise idle GPU cycles to be converted into future cache hits.
Proactive prefetching is beneficial only when execution is sufficiently predictable. \sys therefore continuously evaluates the execution predictability metric $R$ (Eq.~\ref{eq:predictability}). When
\begin{equation}
\label{eq:gate}
R \ge R_{\min},
\end{equation}
the coordinator enables background warmup; otherwise, it disables prefetching and falls back to survival-guided cache replacement alone. This adaptive gating prevents unnecessary GPU work under highly dynamic execution while preserving the benefits of proactive cache management whenever meaningful execution structure exists.

\section{Implementation}
\label{sec:impl}
We implement \sys as a lightweight runtime layer on top of vLLM v0.11 (V1 engine), consisting of approximately 2,300 lines of runtime logic and an 800-line patch. The runtime exposes four primitives, ObserveTouch(), Predict(), ScoreBlock(), and Warmup(), that connect execution-aware cache management to the serving engine.

The patch modifies five vLLM source files (the block pool, the engine core, the scheduler, and the scheduler- and worker-side halves of the CPU-offload connector) and introduces three runtime hooks. First, the block pool invokes ObserveTouch() at prefix-matching time: each request is fingerprinted from its own prefix block hashes, so a fingerprint change signals an agent dispatch and updates transition statistics and the block-to-agent mapping without any framework cooperation (step {\Large\ding{172}} in Algorithm~\ref{alg:runtime}). Second, the eviction path invokes ScoreBlock() (Eq.~\ref{eq:score}) to replace the default LRU ordering with survival-guided ranking, additionally preferring blocks whose contents are already resident in the CPU tier; survival probabilities are computed once per scheduler step and cached. Third, Warmup() is executed between agent turns through the standard serving API: after applying the $R \ge R_{\min}$ gate and a rate limit, it issues an inference request containing only the predicted agent's anchor with \texttt{max\_tokens=1}, which the engine recognizes and excludes from transition learning.

\sys maintains four runtime data structures: an online transition matrix, a sparse execution graph with shortest-path distances, a hop-distance table, and a block-to-agent mapping, totaling under 25\,KB (Sec.~\ref{sec:eval}). Transition counts are updated incrementally on each agent dispatch; the execution graph and hop-distance table are refreshed only when the current agent changes. The eviction hot path performs only a single cached score evaluation (Eq.~\ref{eq:score}); graph construction and background warmup execute entirely off the critical path.

The implementation reuses vLLM's existing execution pipeline without modifying attention kernels, memory allocators, schedulers, or block hashing. Background warmup is issued as a standard inference request, producing the same KV blocks as a normal invocation. \sys can be enabled or disabled through a single environment variable, and reuses vLLM's existing LMCache and CPU offloading infrastructure without modification.

\section{Evaluation}
\label{sec:eval}

This section evaluates the effectiveness and efficiency of \sys on representative multi-agent workloads. Specifically, we seek to answer five questions: (i) Does \sys improve KV-cache reuse and end-to-end serving performance compared with existing approaches? (ii) How much do survival-guided eviction and proactive prefetch contribute to the overall performance gains? (iii) Does \sys generalize to larger models? (iv) Does \sys remain effective under different cache capacities and serving conditions? (v) What runtime overhead does \sys introduce?

\begin{figure}[t]
  \centering
  \vbox to 0pt{\vss\includegraphics[width=0.55\columnwidth]{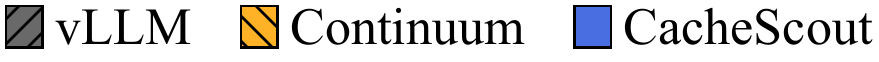}\vss}
  \vspace{0.1in}
  \begin{subfigure}[t]{0.48\columnwidth}
    \centering
    \includegraphics[width=\linewidth]{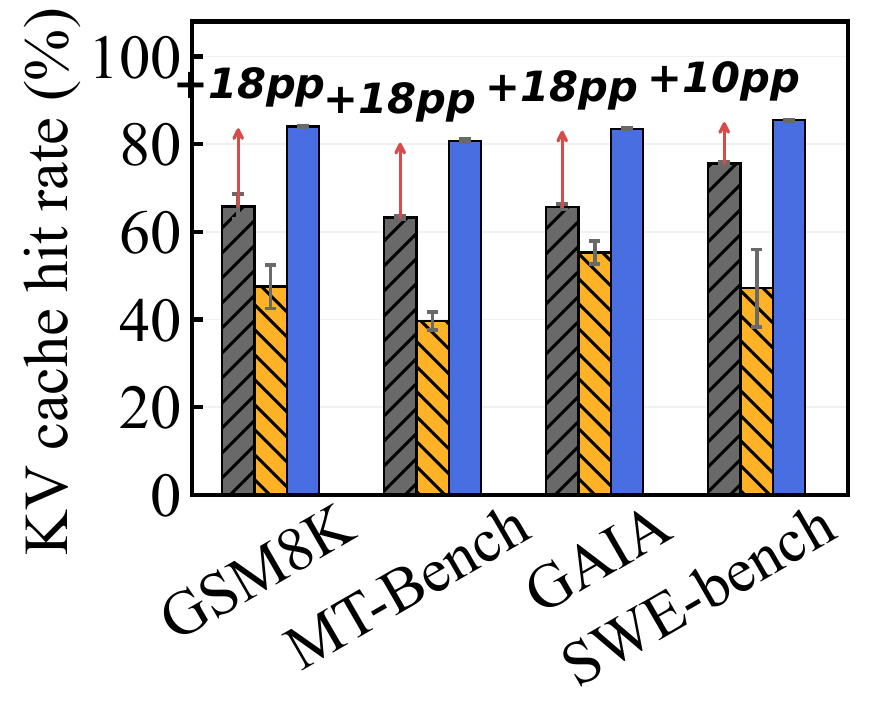}
    \caption{KV-cache hit rate.}
    \label{fig:hit-rate}
  \end{subfigure}
  \hfill
  \begin{subfigure}[t]{0.48\columnwidth}
    \centering
    \includegraphics[width=\linewidth]{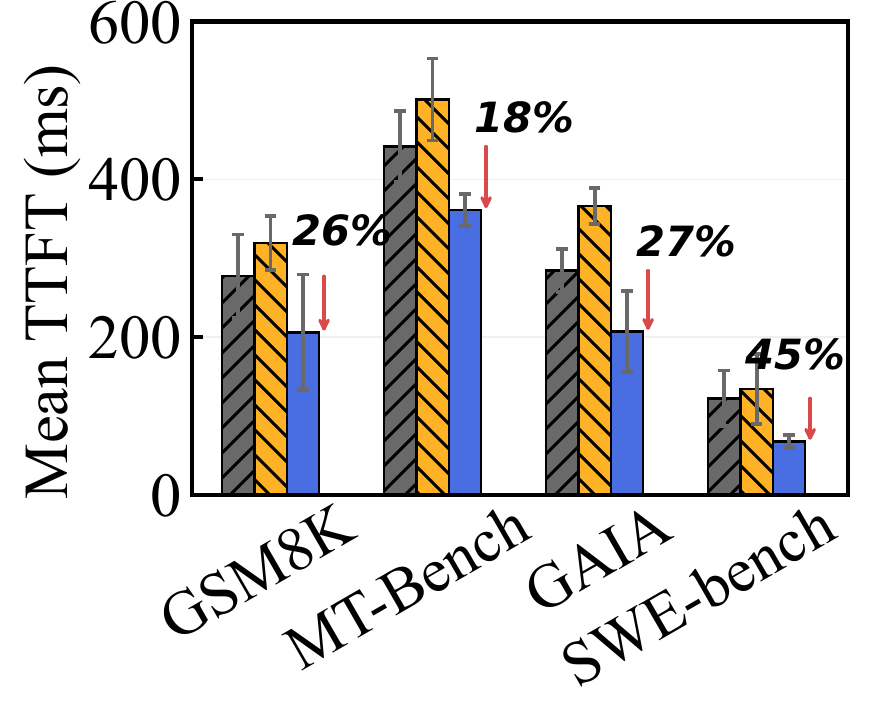}
    \caption{Mean TTFT.}
    \label{fig:latency}
  \end{subfigure}
  \vspace{-0.1in}
  \caption{Higher KV-cache hit rates translate into lower inference latency across representative multi-agent workloads.}
  \label{fig:main-result}
\end{figure}

\begin{figure*}[t]
  \centering
  \vbox to 0pt{\vss\includegraphics[width=0.6\columnwidth]{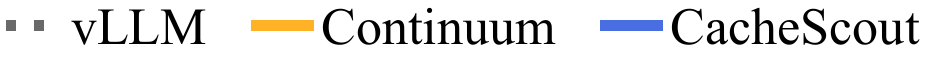}\vss}
  \vspace{0.1in}
  \begin{subfigure}[b]{0.24\textwidth}
    \includegraphics[width=\textwidth]{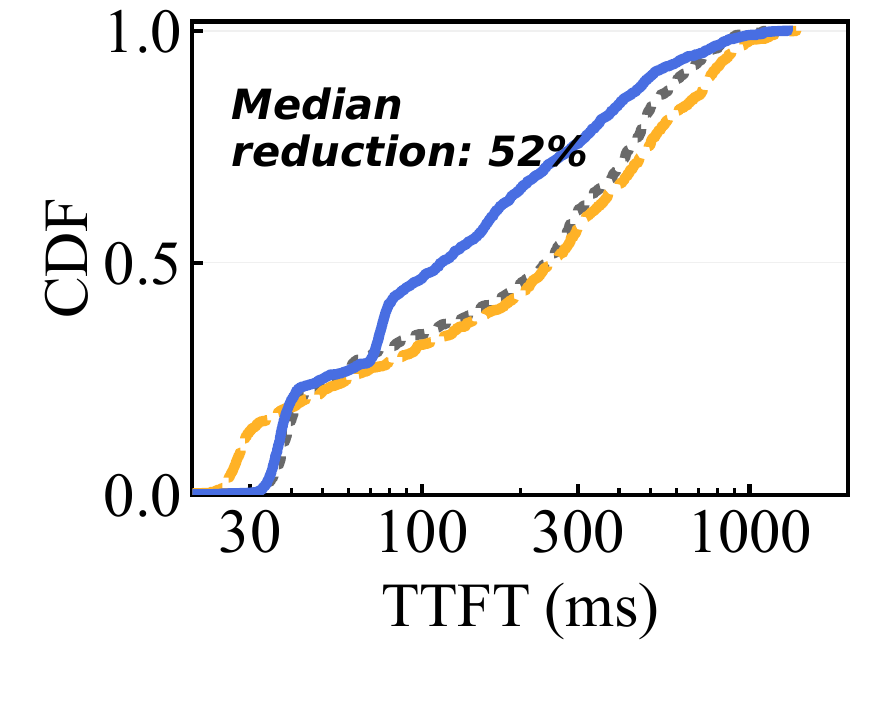}
    \caption{GSM8K}
    \label{fig:ttft_cdf_gsm8k}
  \end{subfigure}
  \hfill
  \begin{subfigure}[b]{0.24\textwidth}
    \includegraphics[width=\textwidth]{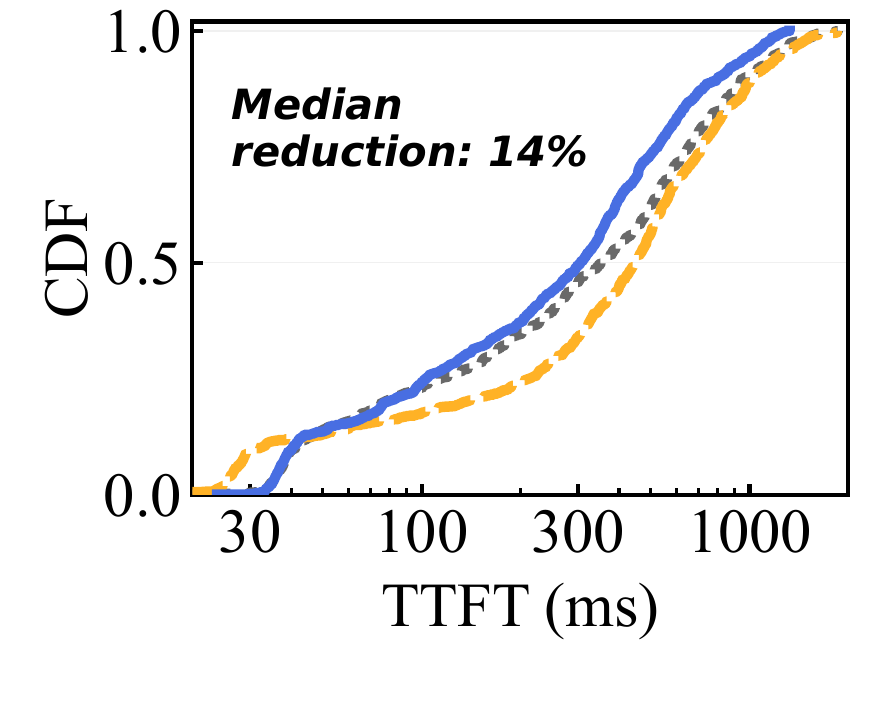}
    \caption{MT-Bench}
    \label{fig:ttft_cdf_mtbench}
  \end{subfigure}
  \hfill
  \begin{subfigure}[b]{0.24\textwidth}
    \includegraphics[width=\textwidth]{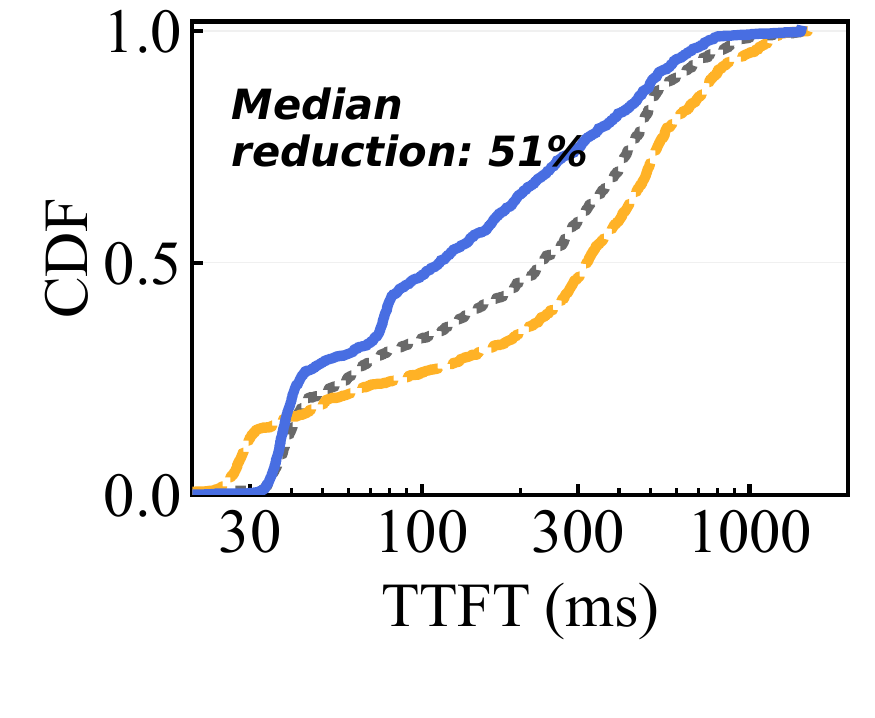}
    \caption{GAIA}
    \label{fig:ttft_cdf_gaia}
  \end{subfigure}
  \hfill
  \begin{subfigure}[b]{0.24\textwidth}
    \includegraphics[width=\textwidth]{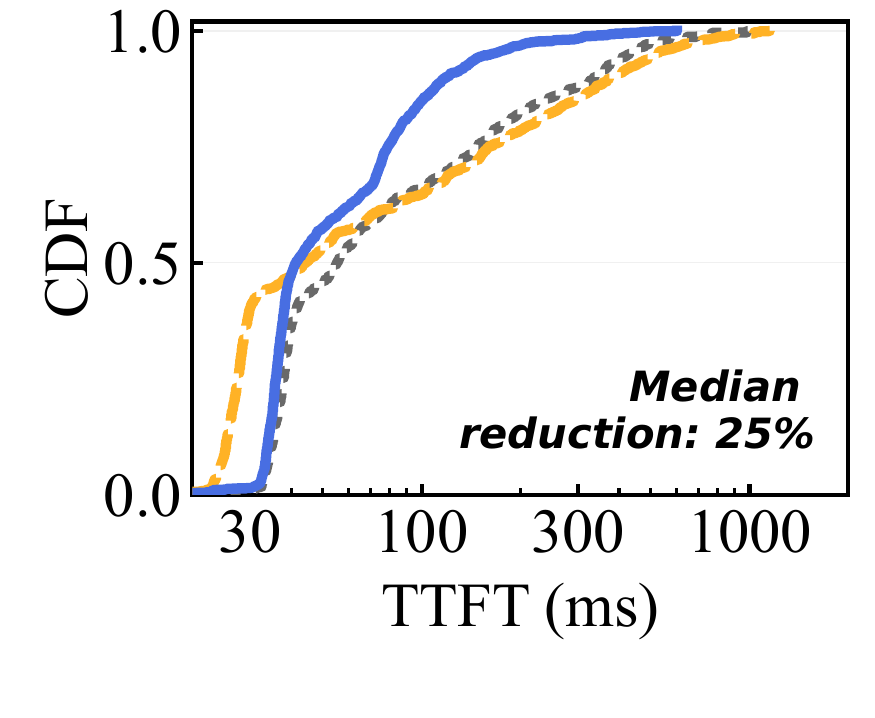}
    \caption{SWE-bench}
    \label{fig:ttft_cdf_swebench}
  \end{subfigure}
  \vspace{-0.1in}
  \caption{TTFT cumulative distributions across representative multi-agent workloads. \sys consistently shifts the distribution toward lower latency.}
  \label{fig:ttft-tail}
  \vspace{-0.1in}
\end{figure*}

\subsection{Experimental Setup}
\noindentparagraph{\textbf{Testbed.}}
We run experiments on a server with eight NVIDIA RTX PRO 6000 Blackwell GPUs, each with 96\,GB of memory. We use Llama-3.1-8B-Instruct as the main model and run all systems under the same vLLM configuration, GPU memory budget, decoding parameters, and prefix cache settings. For the model-scale study (Sec.~\ref{sec:eval:scale}) we use Qwen3-235B-A22B-FP8, a 235B-parameter mixture-of-experts model, with tensor parallelism across four NVIDIA H200 GPUs each with 141\,GB memory and connected via NVLink.

\noindentparagraph{\textbf{Workloads.}} 
We evaluate \sys on four representative multi-agent workloads: GSM8K~\cite{cobbe2021training}, MT-Bench~\cite{zheng2023judging}, GAIA~\cite{mialon2024gaia}, and SWE-bench~\cite{jimenez2024swebench}. All four run on the same six-agent supervisor framework with tools and AutoGen's SelectorGroupChat for dynamic LLM-selected routing. Together, these workloads cover a broad spectrum of execution characteristics, from short mathematical exchanges (GSM8K) and multi-turn conversation (MT-Bench) to tool-heavy web reasoning (GAIA) and long, prefill-heavy software-engineering sessions (SWE-bench).

\noindentparagraph{\textbf{Baselines.}}
We compare \sys against two representative KV-cache management systems. vLLM serves as the production baseline using reactive prefix caching with LRU eviction over block-aligned KV-caches. Continuum represents cross-session KV reuse through TTL-based pinning around tool-call boundaries. Following the observed inter-turn latency in AutoGen, we set the TTL to 0.3\,s. The model-scale study (Sec.~\ref{sec:eval:scale}) compares against vanilla vLLM only, as Continuum's vLLM fork does not support this model.

\noindentparagraph{\textbf{Metrics.}} 
Our evaluation focuses on the following metrics:
\begin{itemize}[leftmargin=*, itemsep=0.4ex, topsep=0.5ex]
    \item \textit{KV-cache hit rate:} The fraction of prompt tokens served from the KV-cache, computed as total\_cached\_tokens / total\_prompt\_tokens. 
    \item \textit{Time-to-first-token (TTFT):} The latency from request arrival to the first generated token, measuring user-perceived responsiveness.
    \item \textit{Per-turn latency:} The end-to-end latency of an individual agent invocation, including both prefill and decoding.
    \item \textit{Throughput:} The number of completed agent turns per second under a fixed workload and hardware configuration.
\end{itemize}

\begin{figure}[t]
  \centering
  \vbox to 0pt{\vss\includegraphics[width=0.5\columnwidth]{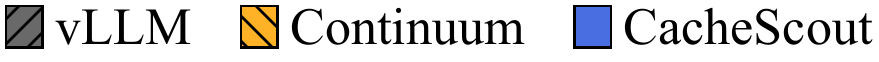}\vss}
  \vspace{0.1in}
  \begin{subfigure}[t]{0.48\columnwidth}
    \centering
    \includegraphics[width=\linewidth]{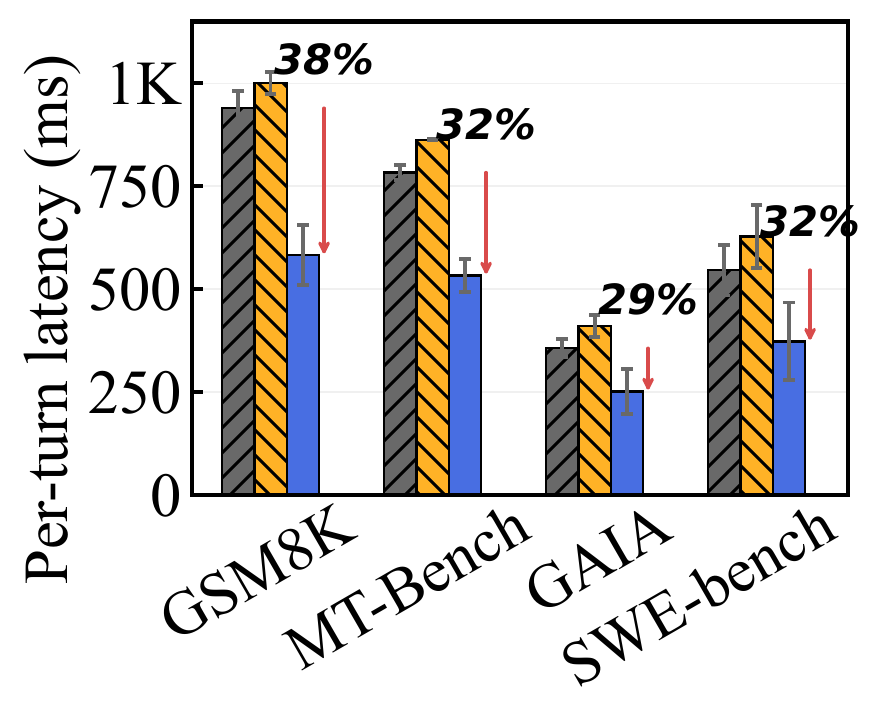}
    \caption{Mean per-turn latency.}
    \label{fig:mean_per_turn_latency}
  \end{subfigure}
  \hfill
  \begin{subfigure}[t]{0.48\columnwidth}
    \centering
    \includegraphics[width=\linewidth]{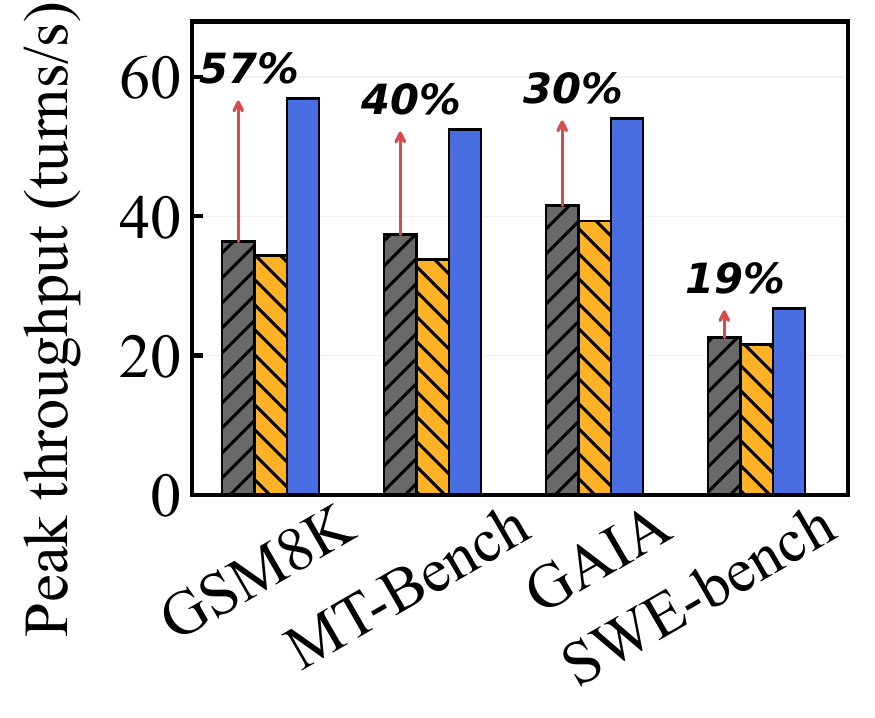}
    \caption{Peak throughput.}
    \label{fig:peak_throughput}
  \end{subfigure}
  \vspace{-0.1in}
  \caption{\sys consistently achieves lower mean per-turn latency and higher peak throughput across representative multi-agent workloads.}
  \label{fig:throughput}
\end{figure}

\subsection{Main Results: \sys improves KV-cache reuse and serving performance}

\sys aims to improve both KV-cache reuse and end-to-end serving performance for dynamic multi-agent workloads. We first examine cache effectiveness through KV-cache hit rate, and then evaluate its impact on serving performance.

\noindentparagraph{\textbf{KV-cache hit rate and Time-to-first-token.}}
Figure~\ref{fig:hit-rate} shows that \sys consistently achieves the highest KV-cache hit rate across all four workloads, improving over vanilla vLLM by 10--18 percentage points despite dynamic LLM-selected routing. The higher hit rate translates directly into faster responses: Figure~\ref{fig:latency} shows that \sys reduces mean TTFT by 18--45\% across all four workloads. Figure~\ref{fig:ttft-tail} further shows that the improvement holds across the latency distribution: median TTFT drops by 14--52\% (e.g., 231${\to}$114\,ms on GAIA and 239${\to}$115\,ms on GSM8K), 
and P99 TTFT drops by 21--52\% on three of the four workloads (711${\to}$342\,ms on SWE-bench),
so \sys is below vanilla vLLM at the median on every workload. This indicates that the benefit is not limited to a small fraction of requests but consistently improves user-perceived responsiveness.

\begin{figure*}[t]
  \centering
  \vbox to 0pt{\vss\includegraphics[width=0.6\columnwidth]{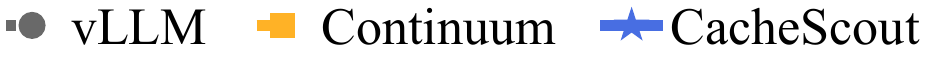}\vss}
  \vspace{0.1in}
  \includegraphics[width=\linewidth]{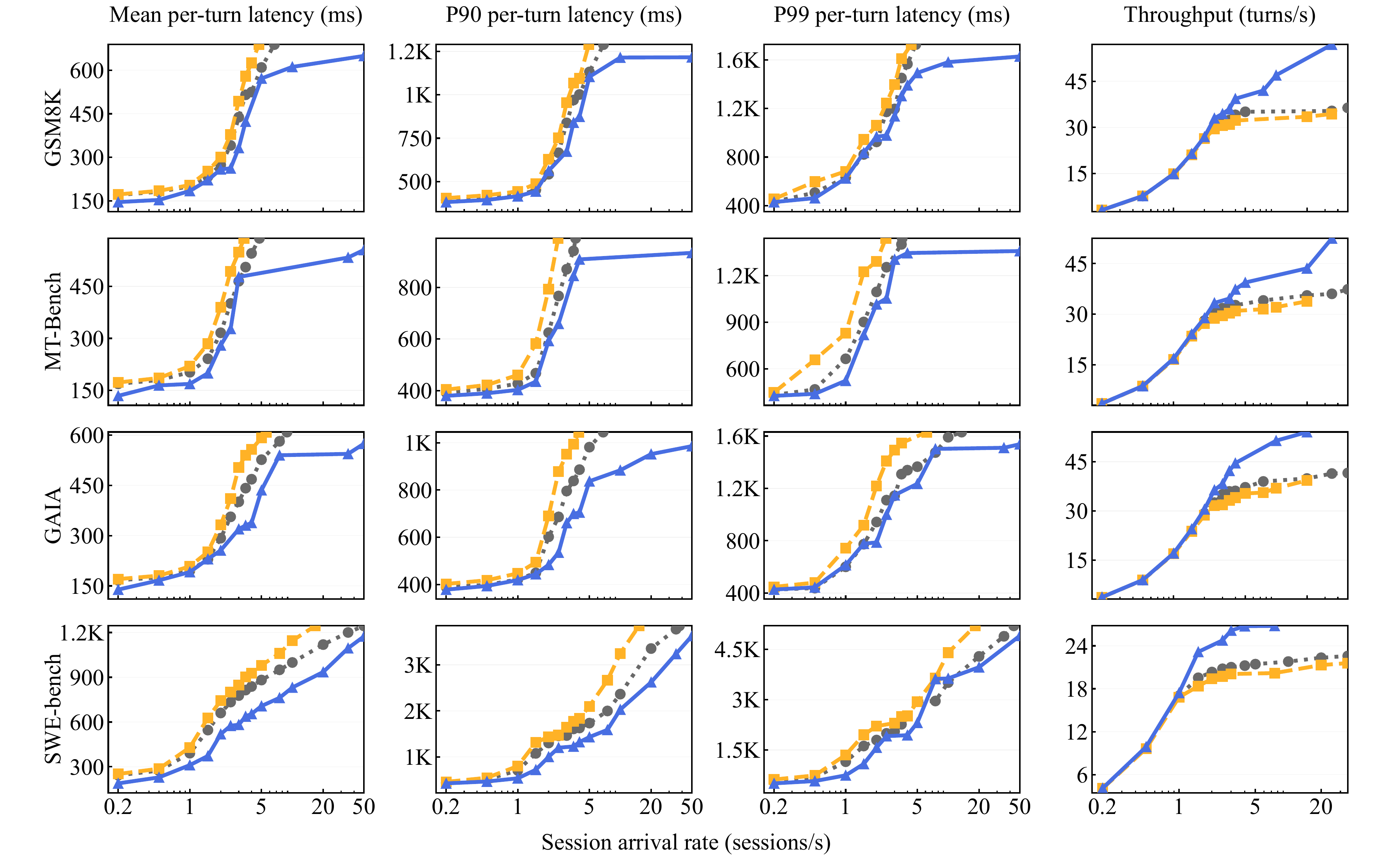}
  \vspace{-0.3in}
  \caption{Latency and throughput versus session arrival rate. \sys delays cache saturation, sustaining lower latency and higher throughput under increasing load.}
  \label{fig:qps-grid}
\end{figure*}

\noindentparagraph{\textbf{End-to-end latency and throughput.}} 
The higher KV-cache hit rate translates directly into better serving performance. As shown in Figure~\ref{fig:throughput}, \sys reduces mean per-turn latency by 29--38\% while delivering 19--57\% higher peak throughput across all four workloads. Figure~\ref{fig:qps-grid} shows how this advantage compounds with load: vanilla vLLM's throughput plateaus once its KV-cache saturates, while \sys continues to scale, and within the same mean-latency budget \sys sustains $1.7$--$12\times$ the arrival rate of vanilla vLLM and $4.2$--$16\times$ that of Continuum. Together, these results demonstrate that improving KV-cache reuse benefits not only request latency but also the overall efficiency of multi-agent serving.

\subsection{Design Breakdown: Predictive eviction provides the primary performance gains
}

\begin{figure*}[t]
  \centering
  \vbox to 0pt{\vss\includegraphics[width=0.73\columnwidth]{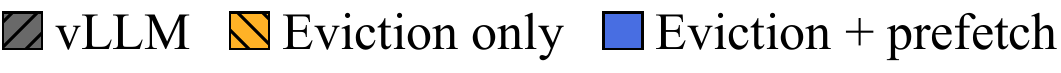}\vss}
  \vspace{0.1in}
  \begin{subfigure}[b]{0.24\textwidth}
    \includegraphics[width=\textwidth]{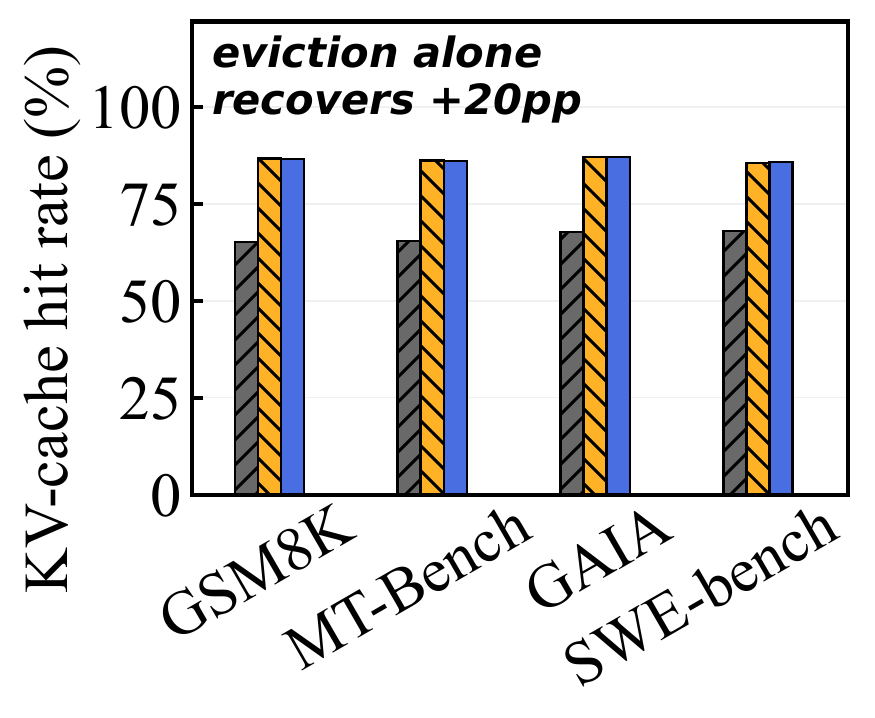}
    \caption{KV-cache hit rate.}
    \label{fig:a1_hit_rate}
  \end{subfigure}
  \hfill
  \begin{subfigure}[b]{0.24\textwidth}
    \includegraphics[width=\textwidth]{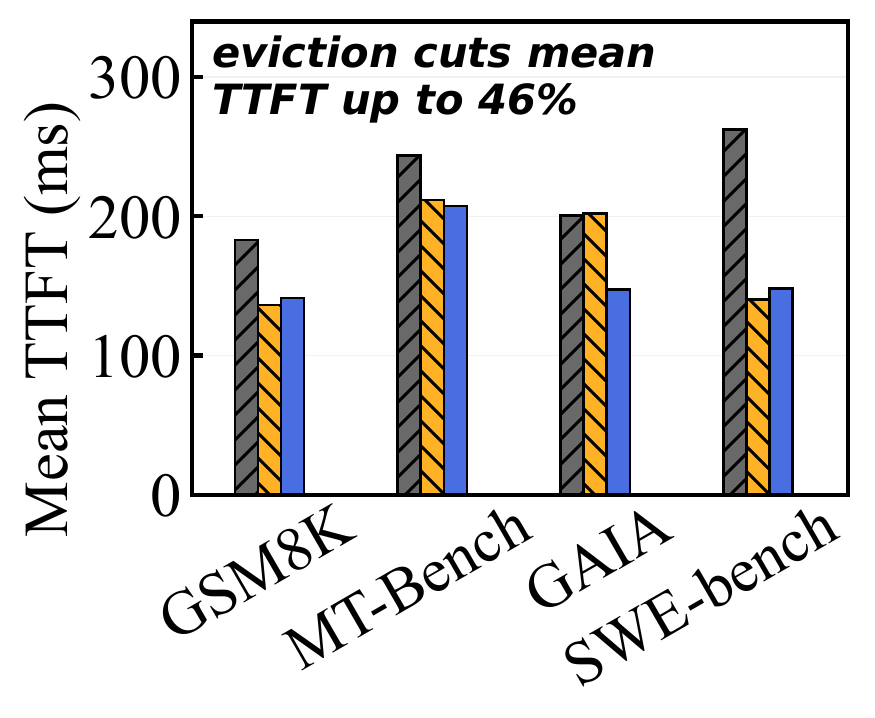}
    \caption{Mean TTFT.}
    \label{fig:a1_mean_ttft}
  \end{subfigure}
  \hfill
  \begin{subfigure}[b]{0.24\textwidth}
    \includegraphics[width=\textwidth]{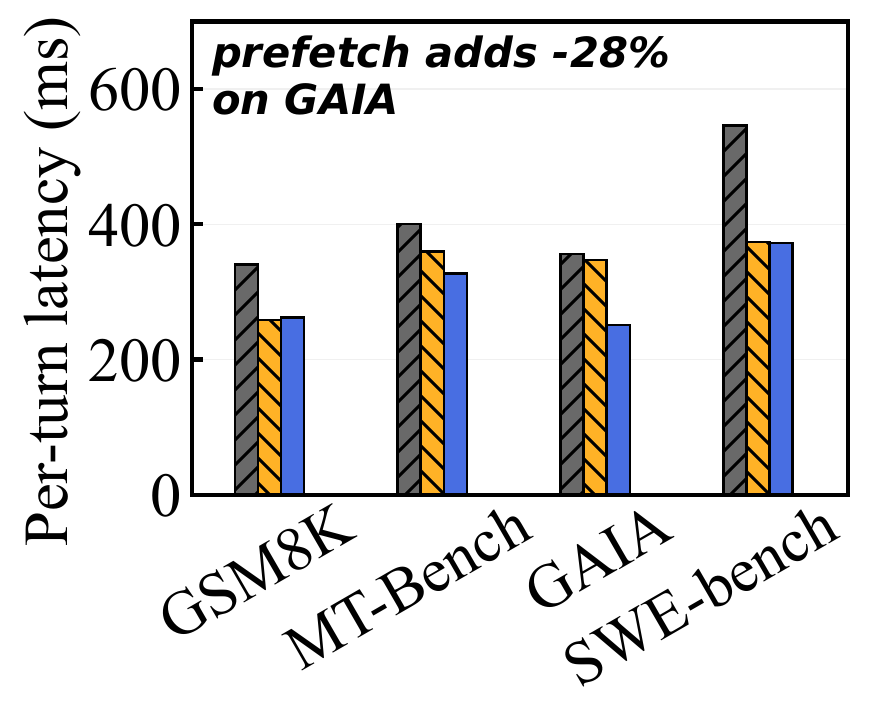}
    \caption{Mean per-turn latency.}
    \label{fig:a1_mean_per_turn_latency}
  \end{subfigure}
  \hfill
  \begin{subfigure}[b]{0.24\textwidth}
    \includegraphics[width=\textwidth]{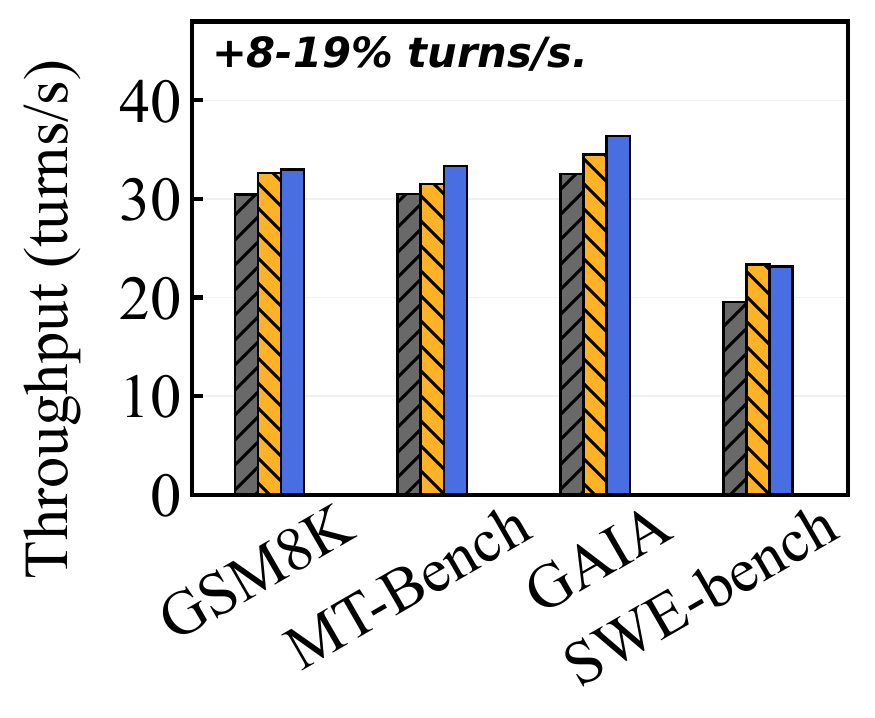}
    \caption{Peak throughput.}
    \label{fig:a1_throughput}
  \end{subfigure}
  \vspace{-0.1in}
  \caption{Mechanism ablation. Survival-guided eviction accounts for most of the cache hit rate improvement, while background prefetch further reduces latency and improves throughput.}
  \label{fig:ablation}
  \vspace{-0.1in}
\end{figure*}


\sys combines two predictive runtime mechanisms: predictive eviction and background prefetch. To understand their individual contributions, we evaluate three configurations: vanilla vLLM, eviction only, and the complete \sys runtime. Figure~\ref{fig:ablation} summarizes their impact on KV-cache hit rate, mean TTFT, per-turn latency, and throughput.

\noindentparagraph{\textbf{Predictive eviction.}}
Figure~\ref{fig:ablation} shows that predictive eviction provides the dominant source of improvement on the four real-world workloads. Enabling predictive eviction alone increases KV-cache hit rate by 18--22 percentage points over vanilla vLLM, whereas background prefetch alone adds at most one percentage point. The higher cache hit rate directly translates into lower per-turn latency and higher throughput (Figures~\ref{fig:a1_mean_per_turn_latency} and~\ref{fig:a1_throughput}), demonstrating that execution-aware cache replacement is the primary contributor to \sys's performance under realistic multi-agent workloads.

\noindentparagraph{\textbf{Background prefetch.}} 
The effectiveness of background prefetch depends on what standalone warmup can preserve. On its own, prefetch is inert on these workloads: the median turn carries no warmable agent prefix, and the blocks a warmup recomputes are evicted by LRU before the predicted agent arrives. Combined with predictive eviction, however, warmup re-activates agent anchors that eviction then protects: the complete runtime reduces per-turn latency by a further 28\% over eviction alone on GAIA (251 vs.\ 347\,ms) and achieves the best overall serving performance on both serving metrics (Figures~\ref{fig:a1_mean_per_turn_latency} and~\ref{fig:a1_throughput}).

These results validate the design of \sys. Predictive eviction serves as the primary optimization for dynamic multi-agent workloads, while background prefetch provides additional gains when execution exhibits strong temporal regularity. This complementary behavior motivates \sys's adaptive prefetch policy.

\subsection{Scalability: \sys remains effective under varying deployment settings and model scale}
\label{sec:eval:scale}

We evaluate the robustness of \sys under varying model scale, GPU cache capacities, and offered load.

\begin{figure*}[t]
  \centering
  \vbox to 0pt{\vss\includegraphics[width=0.45\columnwidth]{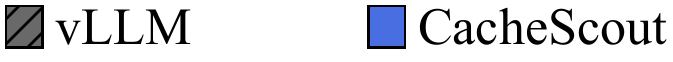}\vss}
  \vspace{0.1in}
  \begin{subfigure}[b]{0.24\textwidth}
    \includegraphics[width=\textwidth]{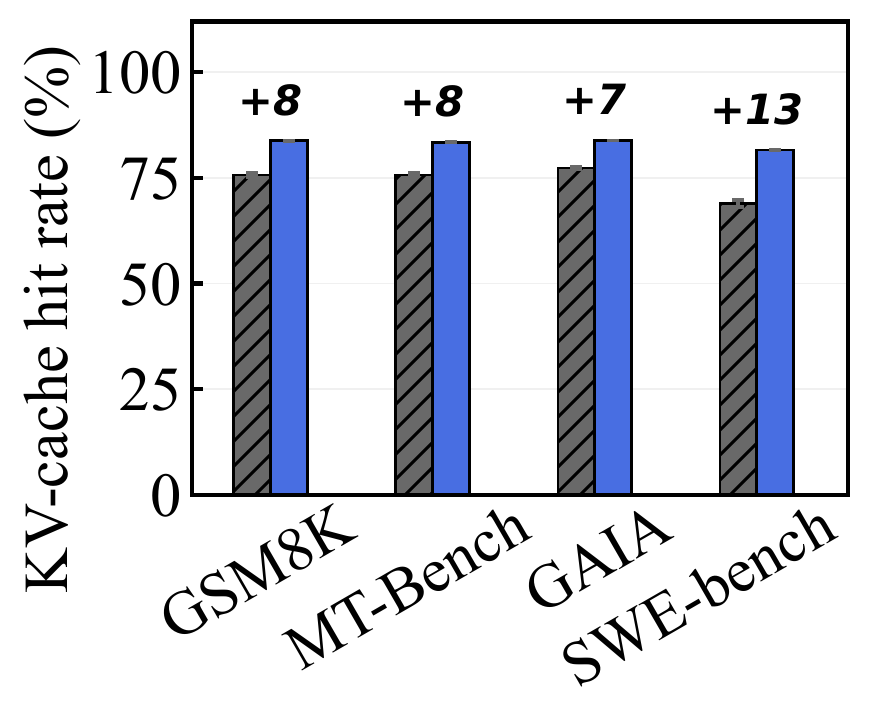}
    \caption{KV-cache hit rate.}
    \label{fig:qwen235_kv_cache_hit_rate}
  \end{subfigure}
  \hfill
  \begin{subfigure}[b]{0.24\textwidth}
    \includegraphics[width=\textwidth]{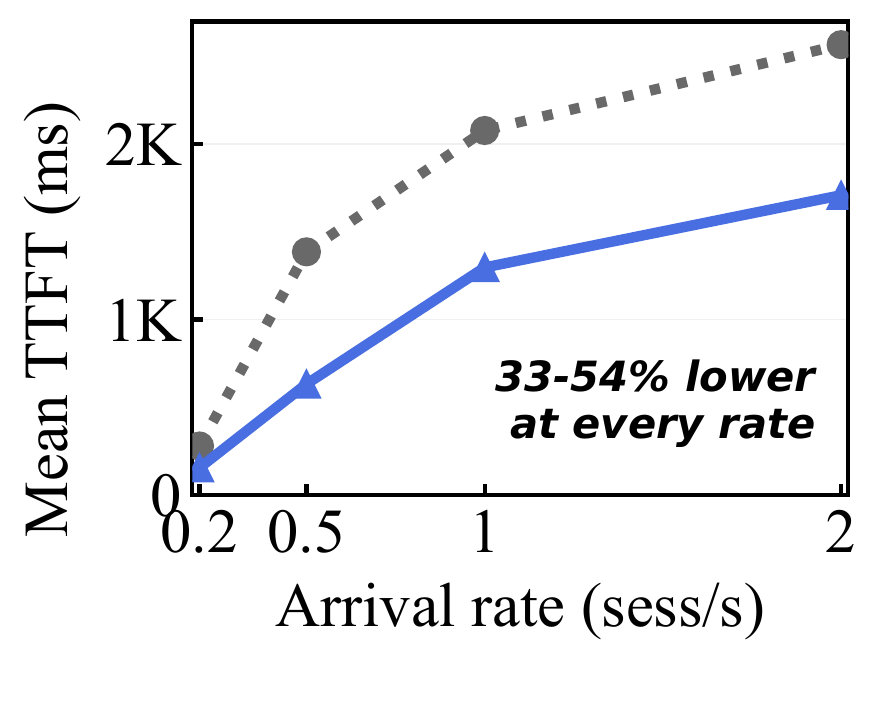}
    \caption{Mean TTFT.}
    \label{fig:qwen235_mean_ttft}
  \end{subfigure}
  \hfill
  \begin{subfigure}[b]{0.24\textwidth}
    \includegraphics[width=\textwidth]{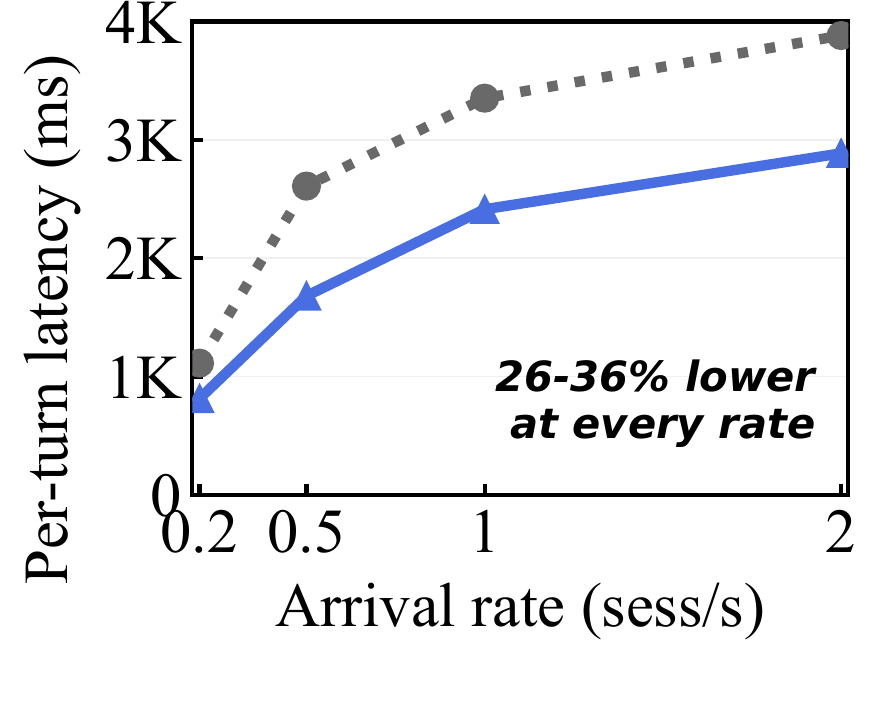}
    \caption{Mean per-turn latency.}
    \label{fig:qwen235_mean_per_turn_latency}
  \end{subfigure}
  \hfill
  \begin{subfigure}[b]{0.24\textwidth}
    \includegraphics[width=\textwidth]{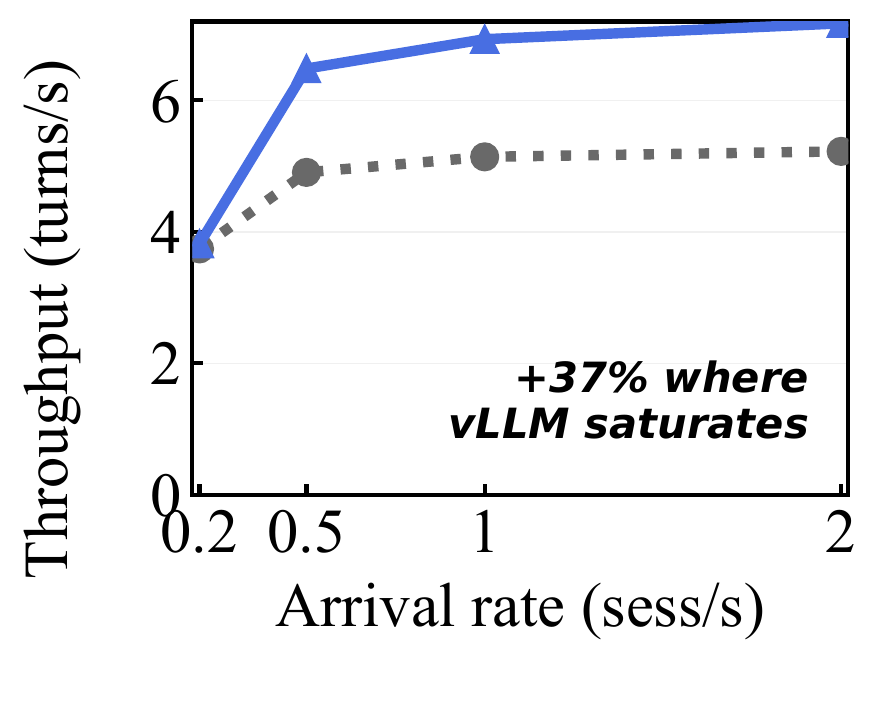}
    \caption{Throughput.}
    \label{fig:qwen235_throughput}
  \end{subfigure}
  \vspace{-0.1in}
  \caption{\sys on Qwen3-235B-A22B-FP8 (tensor parallelism over four H200 GPUs). KV-cache hit rates are reported across all workloads, while latency and throughput are evaluated on SWE-bench. \sys consistently improves cache reuse, reduces inference latency, and sustains higher throughput under increasing load.}
  \label{fig:scale}
\end{figure*}


\begin{figure}[t]
  \centering
  \vbox to 0pt{\vss\includegraphics[width=0.8\columnwidth]{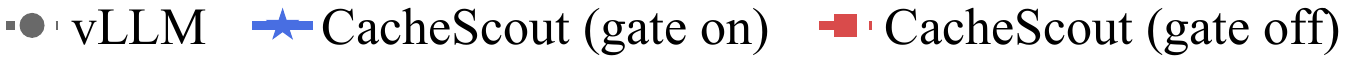}\vss}
  \vspace{0.1in}
  \begin{subfigure}[t]{0.48\columnwidth}
    \centering
    \includegraphics[width=\linewidth]{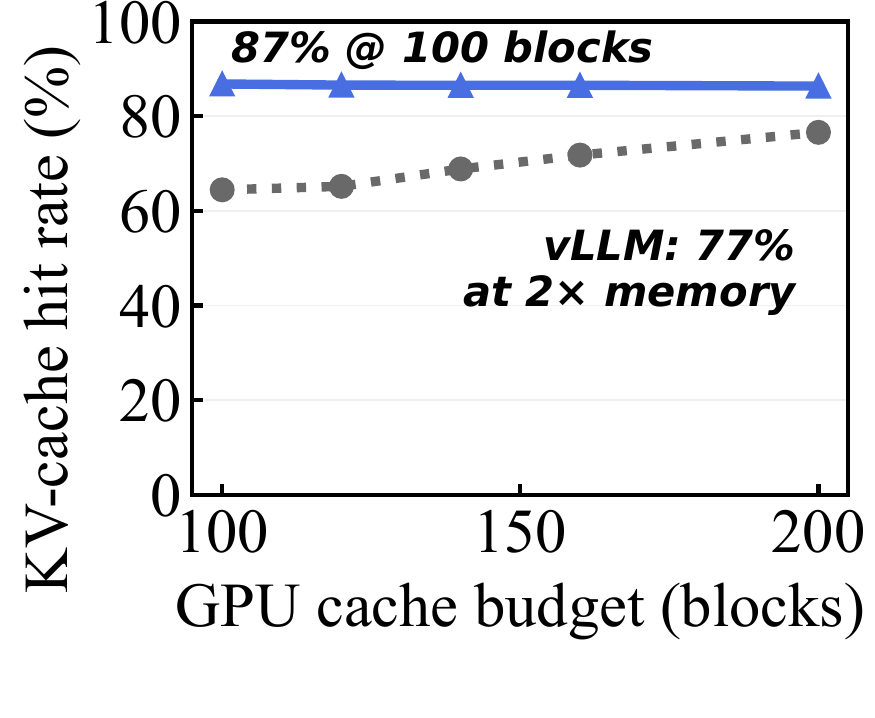}
    \caption{KV-cache hit rate versus GPU cache budget.}
    \label{fig:s_cache_blocks_hit_rate}
  \end{subfigure}
  \hfill
  \begin{subfigure}[t]{0.48\columnwidth}
    \centering
    \includegraphics[width=\linewidth]{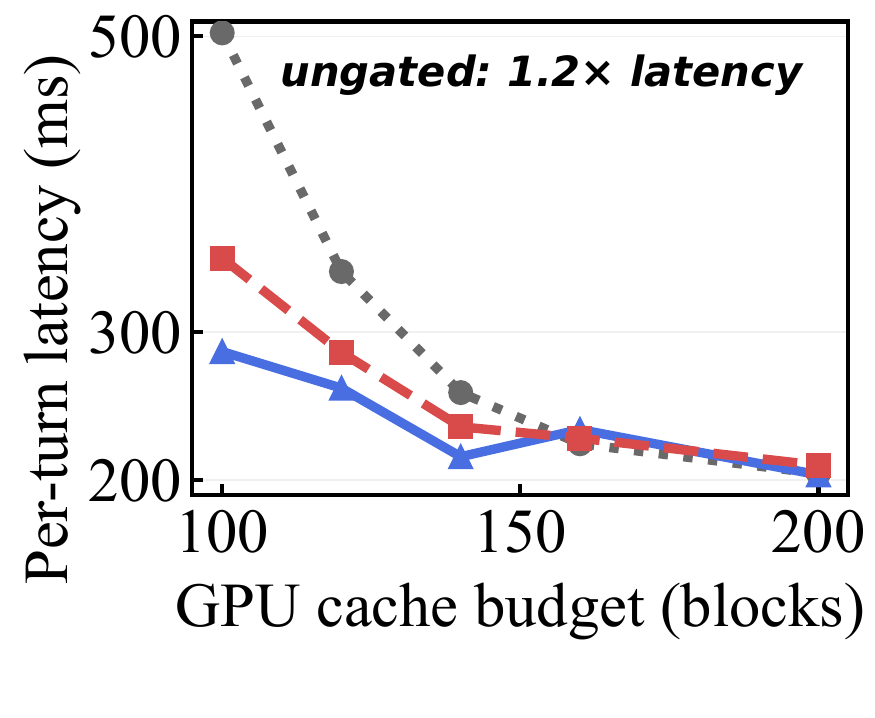}
    \caption{Per-turn latency with adaptive prefetch gating.}
    \label{fig:s_cache_blocks_per_turn_latency}
  \end{subfigure}
  \vspace{-0.1in}
  \caption{Sensitivity to GPU cache budget. \sys maintains high KV-cache hit rates under tight GPU memory budgets, while adaptive prefetch reduces latency without introducing unnecessary warmup overhead.}
  \label{fig:sensitivity}
\end{figure}

\noindentparagraph{\textbf{Model scale.}}
To validate that \sys's mechanisms are not specific to small models, we repeat the serving comparison with Qwen3-235B-A22B-FP8, a 235B-parameter mixture-of-experts model served with tensor parallelism across four H200 GPUs. As shown in Figure~\ref{fig:qwen235_kv_cache_hit_rate}, \sys's hit rate advantage transfers unchanged: it improves KV-cache hit rate by 7--13 percentage points across all four workloads. On SWE-bench, whose long prompts keep prefill on the costly path, this hit rate advantage compounds across the entire load range: sweeping the arrival rate from 0.2 to 2.0 sessions/s (Figure~\ref{fig:scale}b--d), \sys reduces mean TTFT by 33--54\% and mean per-turn latency by 26--36\% at every rate, and where vanilla vLLM's throughput saturates at 5.2 turns/s, \sys continues scaling to 7.2 turns/s ($+$37\%). These results indicate that \sys generalizes across model scale.

\noindentparagraph{\textbf{GPU cache capacity and load.}}
As shown in Figure~\ref{fig:s_cache_blocks_hit_rate}, \sys maintains a nearly constant KV-cache hit rate of 86--87\% as the GPU cache budget varies from 100 to 200 blocks (budgets below the workload's ${\sim}$93-block maximum request footprint cannot admit every request and are excluded). In contrast, vanilla vLLM improves gradually from 64.4\% to 76.6\%, requiring substantially more GPU memory to approach the hit rate achieved by \sys. \sys is equally robust to load: across the full arrival-rate sweep of Figure~\ref{fig:qps-grid} (0.2--50 sessions/s), its KV-cache hit rate holds at 86--87\% while vanilla vLLM stays near 65--68\%. These results demonstrate that \sys remains effective under both memory pressure and increasing load.

\noindentparagraph{\textbf{Adaptive prefetch gate.}}
Figure~\ref{fig:s_cache_blocks_per_turn_latency} evaluates the adaptive prefetch gate introduced in Sec.~\ref{sec:design:prefetch}. With the gate enabled, \sys maintains stable per-turn latency once sufficient cache capacity is available and degrades gracefully as GPU memory becomes more constrained. In contrast, disabling the gate increases per-turn latency by up to 22\% at small cache sizes because unnecessary background warmup competes with foreground inference for GPU resources.

\subsection{Runtime Overhead: \sys incurs negligible runtime overhead}

We evaluate the runtime overhead of \sys through in-process microbenchmarks measuring coordinator memory usage and hot-path latency.


\begin{figure}[t]
  \centering
  \vbox to 0pt{\vss\includegraphics[width=\columnwidth]{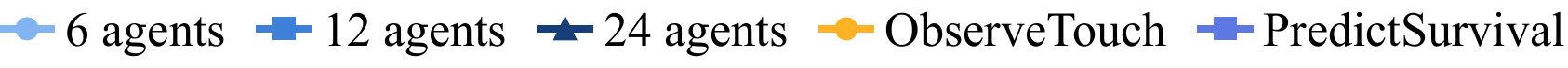}\vss}
  \vspace{0.1in}
  \begin{subfigure}[t]{0.48\columnwidth}
    \centering
    \includegraphics[width=\linewidth]{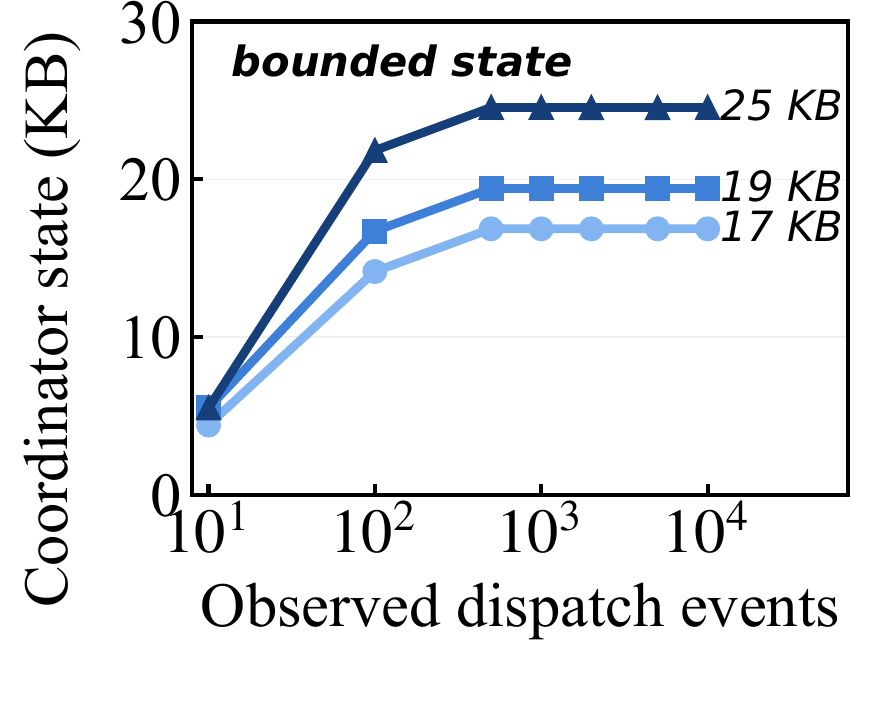}
    \caption{Coordinator state size.}
    \label{fig:o_coordinator_state}
  \end{subfigure}
  \hfill
  \begin{subfigure}[t]{0.48\columnwidth}
    \centering
    \includegraphics[width=\linewidth]{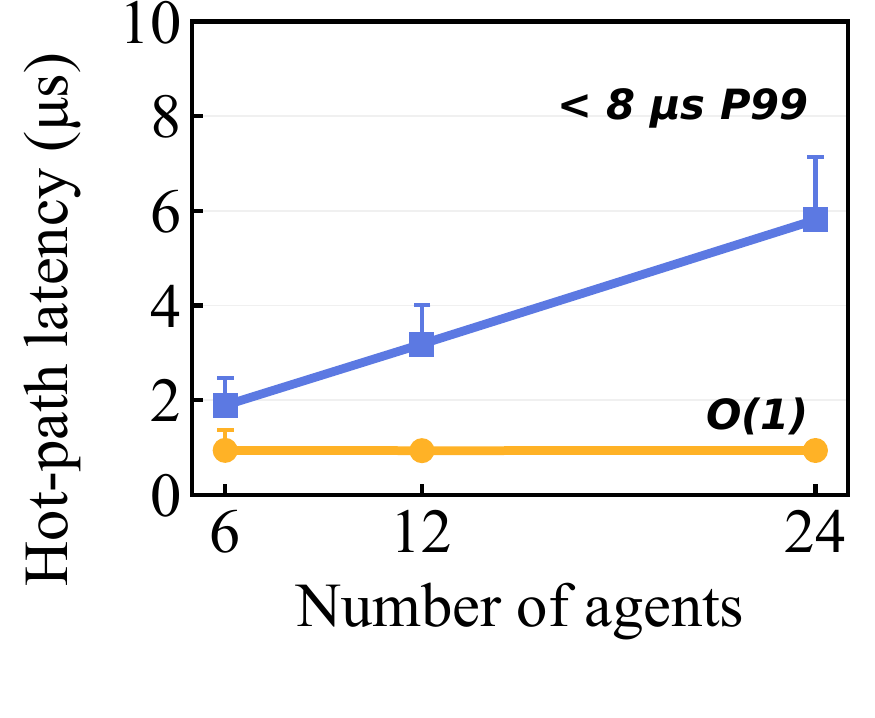}
    \caption{Runtime hot-path latency.}
    \label{fig:o_hot_path_latency}
  \end{subfigure}
  \vspace{-0.1in}
  \caption{Runtime overhead. \sys maintains a compact runtime state and introduces negligible hot-path overhead, even as the number of agents increases.}
\end{figure}

\noindentparagraph{\textbf{Memory footprint.}}
As shown in Figure~\ref{fig:o_coordinator_state}, the coordinator state converges quickly as execution history accumulates and remains below 25 KB even with 24 agents, indicating negligible memory overhead.

\noindentparagraph{\textbf{Runtime latency.}}
Figure~\ref{fig:o_hot_path_latency} shows that both runtime operations incur only microsecond-level overhead. ObserveTouch requires approximately \qty{1}{\us} per block touch, while PredictSurvival completes within \qty{6}{\us} even for 24 agents. These costs are negligible compared with LLM serving latency.


\section{Related Work}
\label{sec:related}

\noindentparagraph{\textbf{DAG-aware KV-cache management.}}
Existing LLM serving engines such as vLLM~\cite{kwon2023vllm} and SGLang~\cite{zheng2024sglang} provide efficient prefix caching, but their cache management remains largely reactive. In agentic workflows, this triggers premature eviction of the agent anchor between steps and forces frequent prefill recompute on bytes the engine just processed.
To solve this, a line of work models the agent calling stack as a DAG known a priori, which lets the runtime anticipate future memory accesses along the declared workflow.
For example, \kvflow~\cite{pan2026kvflow} declares the agent step graph via \sglang \texttt{sgl.function} annotations and scores eviction by step-distance on the declared graph.
SAGA~\cite{guo2026saga} schedules agent workflows atomically using Agent Execution Graphs supplied by the agent framework, and falls back to runtime pattern inference at a $15.6\%$ TCT penalty when framework hints are unavailable.
Parrot~\cite{lin2024parrot} constructs a request DAG from developer-supplied Semantic Variables that annotate input/output placeholders in prompts, and uses the DAG to colocate dependent requests and share common prompt prefixes across them.
PBKV~\cite{zheng2026efficient} also predicts agent invocations to guide eviction and prefetching, but its neural predictor is trained offline per workload and reuses KV only within a single workflow execution.
These systems rely on developer annotations or framework-provided execution graphs, but in real-world agentic workflows such a static DAG often cannot be constructed beforehand. \sys instead learns the per-workload transition matrix online without any deploy-time-declared graph.

\noindentparagraph{\textbf{KV-cache retention across execution stalls.}}
Agentic workloads interleave LLM generation with external function calls; while a tool runs, the GPU pauses, the session's KV is evicted under concurrent traffic, and the next step pays a costly re-prefill when the tool returns.
One line of work prevents the session's KV from being evicted during the tool call, either based on the predicted call duration or on the request's memory footprint over the call's lifetime~\cite{li2025continuum, abhyankar2024infercept, shahout2026fast}.
Another line of work fills the tool pause with other agents' work and reloads the paused agent's KV from CPU just before the tool returns~\cite{bian2025tokencake, luo2025autellix, wei2025agent}.
These approaches retain only the current agent's KV state within a session. \sys addresses a different and broader problem: it prefetches the next agent's anchor at the runtime layer before the call arrives, hitting the cache even when the next call dispatches to a different agent or a different session.

\noindentparagraph{\textbf{Faster agent serving.}}
The latency of individual agents is constrained by the LLM's sequential autoregressive decoding.
Unlike chatbot interactions, agentic workflows are inherently structured, and a line of work accelerates serving by exploiting that structure to bypass or parallelize token-by-token generation.
Token-level methods bypass autoregression via structured generation, constraining output to a grammar so that deterministic spans need not be sampled token-by-token~\cite{dong2025xgrammar, willard2023efficient}.
Sequence-level methods speculatively execute the agent's predicted next step in parallel with the LLM's deliberation~\cite{sui2026act, ro2025sherlock}.
At the framework level, Nalar~\cite{laju2026nalar} decouples workflow specification from execution and schedules agent invocations through dependency-carrying futures.
Runtime layers have also been proposed to enforce serving policies for agentic workloads~\cite{zhang2026policy, zhang2025enabling}.
This line of work is complementary and orthogonal to \sys, and can be combined with \sys to further increase the efficiency of agent inference.

\noindentparagraph{\textbf{Other KV-cache optimizations.}}
Other work optimizes KV-cache management along complementary dimensions, including non-prefix KV-cache sharing with approximation~\cite{yao2025cacheblend, gim2024prompt}, lossy KV-cache reuse with lossless output guarantees~\cite{yao2026vericache}, joint KV-cache compression and eviction~\cite{feng2025evicpress}, cross-LLM KV-cache reusing with partial layer recomputation~\cite{liu2024droidspeak}, request routing that is aware of KV-cache location~\cite{srivatsa2025preble}, and improving paged attention for multi-modality models with different shapes of KV-cache~\cite{tu2025vl}.
While this line of work also makes KV-cache in LLM inference more practical in multiple use cases, \sys is complementary to them as it targets agent-level reuse across session boundaries, an axis orthogonal to these per-block optimizations.

\section{Conclusion}

\sys is an agent-aware KV-cache runtime layer for multi-agent LLM serving that improves cache reuse by leveraging agent execution semantics. 
By learning execution transitions online, it enables predictive cache management without requiring predefined workflow graphs, offline training, or modifications to the serving critical path. 
Experiments on representative real-world multi-agent workloads show that \sys substantially improves KV-cache hit rate, reduces TTFT and end-to-end latency, and increases serving throughput. \sys has been implemented on top of vLLM and will be open-sourced along with the benchmark suite. 
\sys demonstrates that incorporating agent execution semantics into KV-cache management is an effective and practical approach for efficient multi-agent LLM serving.

\balance
\bibliographystyle{ACM-Reference-Format}
\bibliography{references}



\end{document}